\documentclass[preprint,12pt]{elsarticle}

\usepackage{lineno,hyperref,mathrsfs,amssymb,epsfig,graphicx,subfigure,bm,lineno,booktabs,enumerate,booktabs,longtable,array, caption}
\usepackage[labelfont=bf]{caption}
\usepackage[fleqn]{amsmath}
\usepackage{algorithm,algorithmic,multirow,xcolor}
\modulolinenumbers[5]

\journal{Applied Soft Computing}

\begin{document}

\begin{frontmatter}

\title{Multi-view learning with privileged weighted twin support vector machine}


		\author[mymainaddress]{Ruxin Xu}
		\author[mymainaddress]{Huiru Wang \corref{mycorrespondingauthor}}
		\cortext[mycorrespondingauthor]{Corresponding author}
		\ead{whr2019@bjfu.edu.cn}
		
		\address[mymainaddress]{Department of Mathematics, College of Science, Beijing Forestry University, No.35 Qinghua East Road, 100083 Haidian, Beijing, China}
		
\begin{abstract}
Weighted twin support vector machines (WLTSVM) mines as much potential similarity information in samples as possible to improve the common short-coming of non-parallel plane classifiers. Compared with twin support vector machines (TWSVM), it reduces the time complexity by deleting the superfluous constraints using the inter-class K-Nearest Neighbor (KNN). Multi-view learning (MVL) is a newly developing direction of machine learning, which focuses on learning acquiring information from the data indicated by multiple feature sets. In this paper, we propose multi-view learning with privileged weighted twin support vector machines (MPWTSVM). It not only inherits the advantages of WLTSVM but also has its characteristics. Firstly, it enhances generalization ability by mining intra-class information from the same perspective. Secondly, it reduces the redundancy constraints with the help of inter-class information, thus improving the running speed. Most importantly, it can follow both the consensus and the complementarity principle simultaneously as a multi-view classification model. The consensus principle is realized by minimizing the coupling items of the two views in the original objective function. The complementary principle is achieved by establishing privileged information paradigms and MVL. A standard quadratic programming solver is used to solve the problem. Compared with multi-view classification models such as SVM-2K, MVTSVM, MCPK, and PSVM-2V, our model has better accuracy and classification efficiency. Experimental results on 45 binary data sets prove the effectiveness of our method.
\end{abstract}

\begin{keyword}
Multi-view learning \sep Weighted-TWSVM \sep Privileged information \sep Consensus principle \sep Complementarity principle 
\end{keyword}
\end{frontmatter}


\section{Introduction}
Support vector machines (SVMs) \cite{Vapnik1995,Natureofstatisticallearningtheory} are supervised learning models with relevant learning algorithms which is used to analyze data for classification and regression analysis. Till now, many improvements to SVM have been proposed. The multi-surface proximal SVM via generalized eigenvalues (GEPSVM) \cite{GEPSVM} makes each plane the closest to the samples of its category and the farthest away from the samples of other categories. Compared with SVM, the GEPSVM possesses better XOR performance and lower computational complexity. GEPSVM has been developed into a series of new non-parallel plane classifiers. The twin support vector machines (TWSVM) \cite{TSVM2007} has gained wide attention due to its good generalization capacity and short calculation time \cite{Nonparallelplaneproximalclassifier} as a kind of GEPSVM. TWSVM obtains two non-parallel planes by working out two quadratic programming problems (QPPs), and every QPP has a smaller size than the standard SVM.

To further improve the solving speed of TWSVM, Ye et al. proposed a non-parallel plane classifier called weighted TWSVM with local information (WLTSVM) \cite{WLTSVM2012}, which digs as much potential similarity information in the samples as possible. It can discover the geometric and discriminative structures of data manifolds by constructing intra-class and inter-class graphs.
By weighting the samples \cite{Localized_twin_SVM_via_convex_minimization}, the model discovers information about the intrinsic similarity of samples in the same class and derives as many support vectors that reside in the other class as possible.
Based on WLTSVM,
a KNN(K-Nearest Neighbor)-based weighted rough $\nu$-TSVM \cite{KNN-basedroughTWSVM2014}, 
least-squares KNN-based weighted multiclass TSVM \cite{LSKNNTSVM2020},
enhanced regularized KNN-based TSVM (RKNN-TSVM) \cite{RKNN-TSVM2020}
 were proposed so that redundant samples can be deleted  and the running speed can be improved. This idea further reduces the effect of outliers on the model.

Multi-view data are feature data of the same object obtained from different ways or levels, with polymorphism characteristics, multi-source, multi-descriptive, and high-dimensional heterogeneity, etc \cite{MVClustering2014}. For example, SIFT features, color histogram features, texture features, and text descriptions constitute a multi-view of the image in the image recognition problem.
Multi-view data use features distributed in different feature spaces from different perspectives to describe an object. These various features disclose different attributes of objects from distinct perspectives, thus enabling a more comprehensive and accurate description of objects compared to a single perspective.
Multi-view learning (MVL), also considered data integration of multiple feature sets, is a rising direction in machine learning. Nowadays, MVL has been widely used in various fields and researches \cite{multiview1,multiview2,multiview3,multiviewscaling}.
In addition, a multi-view canonical correlation analysis method based on variational graph neural network is proposed \cite{MVL2021}. This method is an advanced model based on multi-view data and adopts multi-view representation learning techniques.
MVL has proven effective in different application scenarios, such as improving image classification, annotation, and retrieval performance \cite{multiview-graph}, financial distress prediction \cite{finacial}, predicting the multiple stages of AD progression \cite{AD}
and mining product adoption intentions from social media
\cite{Predicting_productMVL2021}.
MVL methods can be deduced in following three major classes \cite{Multiview_overview}: co-regularization style algorithms, co-training style algorithms, and margin consistency style algorithms \cite{MVMED2013,MED-2C2016}.

To better mine the information, MVL generally needs to follow two principles: the principle of consistency and the principle of complementarity \cite{multiview,Largemargin_2c}.
The consistency principle aims to maximize the consistency of multiple views. The principle of complementarity indicates that complementary information from multiple views ought to be used to provide a more comprehensive and accurate description of the object. Under these two principles, MVL algorithms can be sorted into co-regularization algorithms and co-training algorithms.
The co-regularization algorithm fuses the regularization term of the discriminant function into the objective function to insure consensus information between different views.
SVM-2K \cite{SVM-2K2006}, multi-view twin SVMs (MVTSVM) \cite{MVTSVM2015}, regularized multi-view least squares SVM (RMVLSSSVM) \cite{RMVLSSVM2018} and multi-view maximum margin of twin spheres SVM \cite{2021MultiMMTSVM} are representative multi-view co-regularization learning algorithms.
The co-training algorithm maximizes the mutual consistency of multiple views by iterations and exchanges complementary information to generate a classifier on each view \cite{co-traing}.
Algorithms that satisfy these two principles tend to have better generalization capability and performance.
Most existing classification models either satisfy the complementarity principle or the consistency principle, but fewer models satisfy both principles like multi-view least squares SVM \cite{MVLSSVM2018}.

Inspired by the above-mentioned theories and conclusions, we propose a novel classification algorithm for MVL called multi-view learning with privileged weighted twin support vector machines (MPWTSVM). It is achieved by solving two QPPs which makes MPWTSVM work faster. The model we propose mines the potential similarity information between different perspectives and categories by using two graphs (intra-class and inter-class graph) to represent intra-class compactness and inter-class separability.
Intuitively, support vectors exist in the closest relationship between samples that sharing different labels. Therefore, by considering possible support vectors, the time complexity can be significantly reduced.
We ensure the consistency principle by minimizing the coupling terms of the two views in the objective function.
Through the establishment of a privileged information paradigm and MVL, the principle of complementarity is realized.
Consequently, the proposed model coordinates all views' information abundantly during the learning procedure and preserves different views' characteristics and inherent similarity information. We use a standard quadratic programming solver in order to seek the solution of MPWTSVM. In addition, numerical experiments are conducted to verify the performance.

In summary, our contributions can be outlined as follows.
\begin{enumerate}[1)]
	\item MPWTSVM uses the weighted idea of WLTSVM and incorporates this ideology into multiple views. It is worth noting that in the two-classification process, we weigh the samples separately under two perspectives and obtain the weights of different types and the same type at each view. With the help of KNN, redundant samples are deleted. Therefore, the operation efficiency is greatly improved. 
	\item Our model can satisfy both the two principles of MVL, namely, the consensus principle and the complementarity principle. The consensus principle is realized by minimizing the coupling items of the two views in the objective function. The complementary principle is achieved by establishing privileged information paradigms from different perspectives and MVL. Therefore, the proposed MPWTSVM has a better classification ability. 
	\item We compare our method with five state-of-the-art algorithms and performed numerical experiments on 45 binary multi-view classification data sets. The results demonstrate that MPWTSVM has better accuracy and efficiency than other similar algorithms.
\end{enumerate}

The remainder of this paper is presented below. Section 2 retrospects related work about WLTSVM and PSVM-2V. Section 3 provides a detailed description of our proposed MPWTSVM, where we derive the method's dual optimization problem and kernel trick. In Section 4, we compare our algorithm with four state-of-the-art algorithms. Section 5 gives the experimental results, and we provide conclusions and future work in Section 6.

\section{Related Works}
In this section, we give a brief review on a single view learning algorithm WLTSVM and a multi-view learning algorithm PSVM-2V.

\subsection{WLTSVM}
In \cite{WLTSVM2012}, a TSVM-based model WLTSVM with local information was proposed.
The WLTSVM uses two graphs, i.e., intra-class and inter-class graph, to describe the tightness within a class and the separability between classes, so as to dig out as much basic similarity information as possible in the samples.
WLTSVM ﬁnds two nonparallel hyperplanes $f(x)$, one for positive class, and the other for negative class:
$$
f_1(x)=w_1^{\top}x_1+b_1,
f_2(x)=w_2^{\top}x_2+b_2,
$$
where $w_1$ and $w_2$ are two nonparallel hyperplanes' weights, $b_1$ and $b_2$ are the biases.
The model classifies the samples according to the hyperplane to which the given sample is close.

Suppose that we have $N_1$ positive training samples $\{x_i, y_i\}, i = 1,2,...,N_1,$ and $N_2$ negative training samples $\{x_i, y_i\}, i = 1,2,...,N_2,$ where $x_i \in \mathbb{R}$ and $y_i$ is the class label.
For any pair of points $(x_i,x_j), (i = 1,2,...,N_1, j = 1,2,...,N_1)$ in positive samples and an arbitrary point $x_l (l = 1,2,...,N_2)$ in negative samples, we define the weight matrices of within-class graph $G_s\ (W_{s,ij}^1)$ and between-class graph $G_d\  (f_j^1)$ of positive samples as below:
\begin{equation}
W_{s,ij}^1=\left\{
\begin{aligned}
1, &\quad \text{if}\ x_i\ \text{is}\ \text{the}\ \text{ k-nearest}\ \text{neighbors}\ \text{of}\ x_j \\
&\quad  \text{or}\ x_j\  \text{is}\  \text{the}\  \text{ k-nearest}\ \text{neighbors}\ \text{of}\ x_i \\
0, &\quad \text{otherwise},
\end{aligned}\right.
\end{equation}
\begin{equation}
f_j^1=\left\{
\begin{aligned}
1, &\quad \exists j, W_{d,ij}^1 \not= 0\\
0, &\quad \text{otherwise},
\end{aligned}\right.
\end{equation}
where
\begin{equation}
W_{d,ij}^1=\left\{
\begin{aligned}
1, &\quad \text{if}\ x_l\ \text{is}\ \text{the}\ \text{ k-nearest}\ \text{neighbors}\ \text{of}\ x_i \\
0, &\quad \text{otherwise}.
\end{aligned}\right.
\end{equation}

Similarly, we can get the weight matrices of negative samples and name them $W_{s,ij}^2$ and $f_j^2$.

Let $d_j=\sum_{j=1}^{l_1} W_{s,ij}^1,j=1,2,...,N_1;q_j=\sum_{j=1}^{l_2} W_{s,ij}^2,j=1,2,...,N_2$,then the formulation of WLTSVM can be written as follows:
\begin{align}
\label{WLTSVM1}\mathbf{\min} \quad
 &\frac{1}{2} \sum_{i=1}^{l_1} d_j(\omega_1^{\top} x_j^+ +b_1)^2 +C \sum_{j=1}^{l_2} \xi_j, \nonumber\\
 s.t.
\quad & -f_j^1(\omega_1^{\top} x_j^- +b_1)+ \xi_j \ge f_j^1 \cdot 1, \quad \xi_j \ge 0,
\end{align}
and
\begin{align}
\label{WLTSVM2}\mathbf{\min} \quad
 &\frac{1}{2} \sum_{i=1}^{l_2} q_j(\omega_2^{\top} x_j^+ +b_2)^2 +C \sum_{j=1}^{l_1} \xi_j, \nonumber\\
 s.t.
\quad & -f_j^2(\omega_2^{\top} x_j^+ +b_2)+ \xi_j \ge f_j^2 \cdot 1, \quad \xi_j \ge 0,
\end{align}
where $(w_i, b_i) \in (R_n \times R) (i = 1,2)$, C is the penalty coefficient, and $\xi$ is the nonnegative slack variable.

\subsection{PSVM-2V}
Suppose there is a dataset with two perspectives of $l$ labeled augmented samples $\{(X_A, X_B, Y)\} =\{(x_i^A, x_i^B, y_i )\}_{i=1}^l=\{(x_i^A;1),(x_i^B;1),y_i)\}_{i=1}^l$, where $X_A$ and $X_B$
are two views’ feature spaces. The $i^{th}$ sample of two views are represented by the superscripts
A and B of $x_i$.
PSVM-2V ﬁnds two hyperplanes, one for view A, another for view B:
$$
{w_A^*}^{\top}  \phi_A(x^A)=0,
{w_B^*}^{\top}  \phi_B(x^B)=0,
$$
with the optima $w_A^*$ and $w_B^*$ from problem (\ref{PSVM-2V}).

The optimization problem of PSVM-2V \cite{PSVM-2V2017} can be written as follows,
\begin{align}
  \mathbf{\min}
  \label{PSVM-2V}
  \quad
  &\frac{1}{2} \left(||\omega_A||^2+\gamma ||\omega_B||^2\right) + C^A \sum_{i=1}^l \xi_i^{A^*} + C^B \sum_{i=1}^l \xi_i^{B^*} + C \sum_{i=1}^l \eta_i \nonumber\\
  s.t.\quad
  &|(w_A \cdot \phi_A(x_i^A))-(w_B \cdot \phi_B(x_i^B))| \le \epsilon+\eta_i \nonumber\\
  & y_i(w_A \cdot \phi_A(x_i^A)) \ge 1-\xi_i^{A^*}  \nonumber\\
  & y_i(w_B \cdot \phi_B(x_i^B)) \ge 1-\xi_i^{B^*}  \nonumber\\
  & \xi_i^{A^*} \ge y_i(w_B \cdot \phi_B(x_i^B)), \quad \xi_i^{A^*}\ge 0\nonumber\\
  & \xi_i^{B^*} \ge y_i(w_A \cdot \phi_A(x_i^A)), \quad  \xi_i^{B^*}\ge 0\nonumber\\
  &\eta_i \ge 0,\quad i=1,...,l,
\end{align}
where $w_A$, $w_B$ are weight vectors for views
A and B respectively and $\phi_A,\phi_B$ are mappings from inputs to high-dimensional feature spaces. The principle of complementary is realized by limiting the non-negative slack variables $\xi_A=(\xi_1^A,\xi_2^A,...\xi_l^A)^{\top}$ and $\xi_B=(\xi_1^B,\xi_2^B,...\xi_l^B)^{\top}$ by the non-negative correction function. $C^A$, $C^B$, $C$ are non-negative penalty parameters, $\gamma$ is a non-negative
trade-off parameter and $\eta$ is the nonnegative slack variable.

\section{The multi-view learning with priviledged weighted twin support vector machines}

In this paper, we propose a novel MVL method called multi-view learning with priviledged weighted TSVM  (MPWTSVM) which realizes the consensus and complementarity principle at the same time. It not only inherits the weighting idea of WLTSVM, but also combines MVL to produce better performance. The linear and nonlinear cases of MPWTSVM and the dual formulations are presented in the following sections. Major notations used in this paper are summerized in Table \ref{notations}.

\begin{longtable}[c]{ll}
\caption{List of notations}\\
\toprule\label{notations}
Notation                      & Description                                             \\ \hline
\endhead
\hline
\endfoot
\endlastfoot
$(x_i^A,x_i^B,y_i)$           & $i$th training point                                    \\
$l$                           & number of training points                               \\
$(x_i \cdot x_j)$             & inner product between $x_i$ and $x_j$ as $x_i^{\top} x_j$    \\
$\omega^A,\omega^B$           & weight vectors for view A and view B                    \\
$\phi (\cdot)$               & mappings from inputs to high-dimensional feature spaces \\
$\mathcal{K}(x_i,x_j)$  & kernel function $(\phi(x_i) \cdot \phi(x_j))$   \\
$C_A,C_B,C,C_{A2},C_{B2},C_2$ & non-negative penalty parameter                                       \\
$\gamma$                      & non-negative trade-off parameter                                     \\
$W^A,W^B$                 & intra-class weight matrix of view A and view B  \\
$f_i^A,f_i^B,f_j^A,f_j^B$  & inter-class weight matrix of view A and view B \\
\bottomrule
\end{longtable}

\subsection{Linear MPWTSVM}
To make full use of similarity information in data affinity we define the intra-class weight matrix of view A's positive and negative samples respectively.
\begin{equation}
W_{s,ij}^A=\left\{
\begin{aligned}
1, &\quad \text{if}\ x_i^A\ \text{is}\ \text{the}\ \text{ k-nearest}\ \text{neighbors}\ \text{of}\ x_j^A \\ &\quad  \text{or}\ x_j^A\  \text{is}\  \text{the}\  \text{ k-nearest}\ \text{neighbors}\ \text{of}\ x_i^A \\
0, &\quad \text{otherwise},
\end{aligned}\right.
\end{equation}
pairs of points $(x_i^A,x_j^A)$ are in view A's positive and negative samples respectively. Similarly, we can acquire intra-class weight matrix of view B's positive samples and negative samples respectively.

The inter-class weight matrix can be defined as below:
\begin{equation}
f_j^A=\left\{
\begin{aligned}
1, &\quad \exists j, W_{d,ij}^A \not= 0\\
0, &\quad \text{otherwise},
\end{aligned}\right.
\end{equation}
\begin{equation}
\text{where} \ W_{d,ij}^A=\left\{
\begin{aligned}
1, &\quad \text{if}\ x_l^A\ \text{is}\ \text{the}\ \text{ k-nearest}\ \text{neighbors}\ \text{of}\ x_i^A \\  0, &\quad \text{otherwise},
\end{aligned}\right.
\end{equation}
and pairs of points $(x_i^A,x_j^A)$ are in view A's positive samples, $x_l^A$ is an arbitrary point in view A's negative samples.
In the same way, the inter-class weight matrix of view A and view B can be well defined.
We can use the weighted thought of WLTSVM and incorporate this idea into multiple views. By this means, we can obtain the intra-class and inter-class weight matrix weight in the corresponding view.

\begin{figure}[ht!]
    \includegraphics[width=\textwidth]{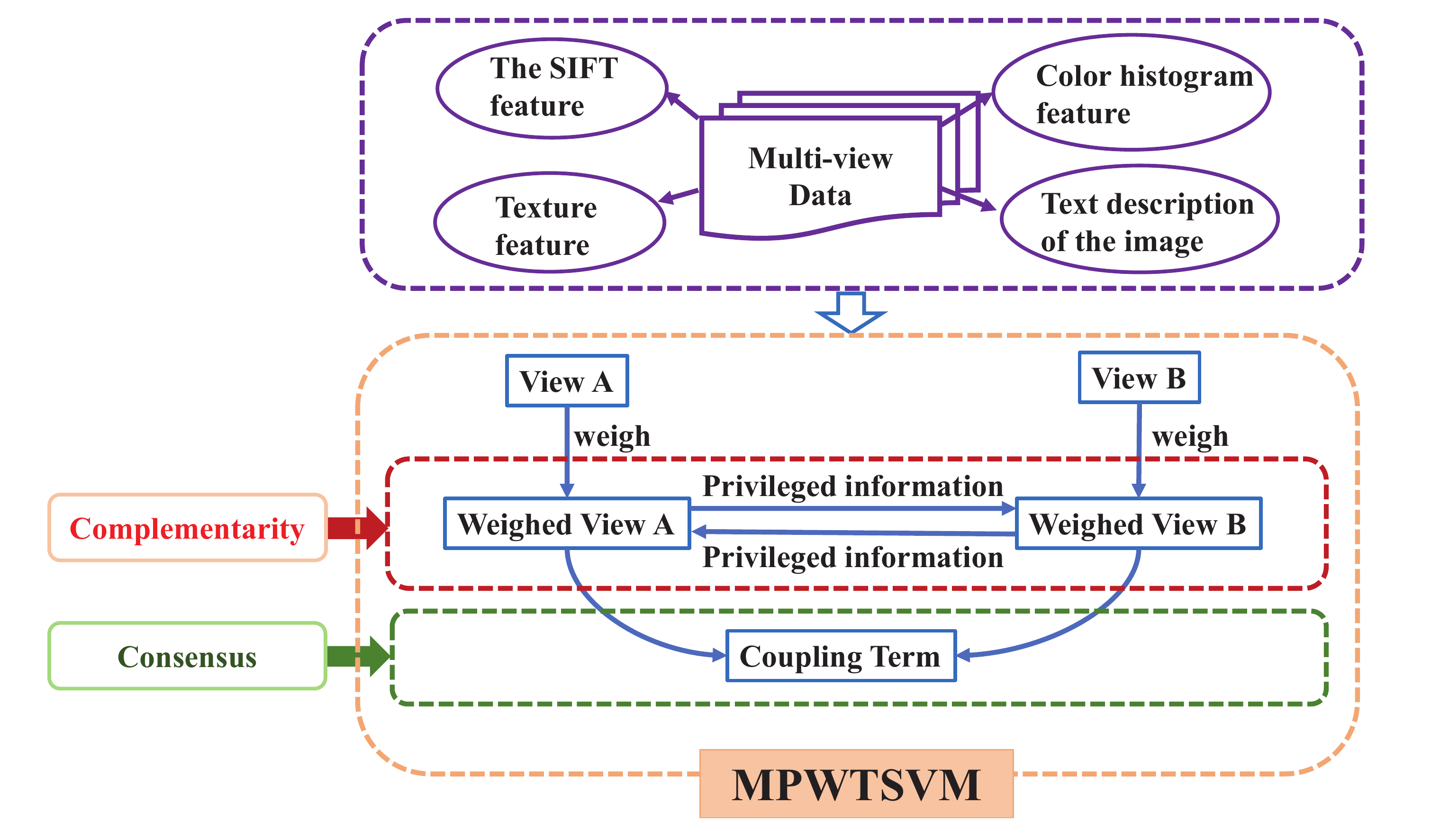}
    \caption{Schematic diagram of MPWTSVM model construction.
}
    \label{lctpic}
    \end{figure}

Fig.\ref{lctpic} displays the model construction of MPWTSVM. Multi-view data are gathered from different fields or gained from different feature extractors for effective learning.
By weighting the data from different perspectives, MPWTSVM makes full use of the similarity information in data affinity. Combining with privilege information and introducing coupling items, MPWTSVM meets the two principles of multi-view classification.

MPWTSVM ﬁnds four hyperplanes, two for view A, two for view B:
\begin{align}
& f_1^A(x^A)=w_1^A x_1^A+b_1^A,\quad
f_2^A(x^A)=w_2^A x_2^A+b_2^A,\nonumber\\
& f_1^B(x^B)=w_1^B x_1^B+b_1^B,\quad
f_2^B(x^B)=w_2^B x_2^B+b_2^B,\nonumber
\end{align}
where $f_1^A(x^A)$ and $f_2^A(x^A)$ are two nonparallel hyperplanes for positive class and negative class of view A  separately,  $f_1^B(x^B)$ and $f_2^B(x^B)$ are two nonparallel hyperplanes for positive class and negative class of view B.
$w_1^t$ and $w_2^t$ are the weights of two nonparallel hyperplanes, $b_1^t$ and $b_2^t$ are the biases of view t (t=A,B).
The  view t's model of MPWTSVM classiﬁes the samples relying on which hyperplane (from $f_1^t(x)$ and $f_2^t(x)$) the given view t's sample is close to. 

Formally, MPWTSVM can be established as follows:
\begin{align}\label{MPWTSVM_Q1}
\mathbf{\min} \quad
 &\frac{1}{2} \sum_{i=1}^{l_1}\sum_{j=1}^{l_1}{W_{s,ij}^{A}} (\omega_{+}^{A^{\top}} x_{j}^{A,+}+b_{+}^{A})^2
 +\frac{1}{2} \gamma \sum_{i=1}^{l_1}\sum_{j=1}^{l_1}{W_{s,ij}^{B}} (\omega_{+}^{B^{\top}} x_{j}^{B,+}+b_{+}^{B})^2  \nonumber\\
 & +C_{A}\sum_{j=1}^{l_2}{\xi_{j}^{A}} +C_{B}\sum_{j=1}^{l_2}{{\xi_{j}^{B}}+C\sum_{j=1}^{l_2}{\xi_{j}^{A}\xi_{j}^{B}}}\nonumber\\
s.t.
\quad & -f_{j}^{A}(\omega_{+}^{A^{\top}} x_{j}^{A,-}+b_{+}^{A})+\xi_{j}^{A} \ge f_{j}^{A} \cdot 1, \nonumber\\
& -f_{j}^{B}(\omega_{+}^{B^{\top}} x_{j}^{B,-}+b_{+}^{B})+\xi_{j}^{B} \ge f_{j}^{B} \cdot 1, \nonumber\\
&\quad \xi_{j}^{A} \ge -f_{j}^{B}(\omega_{+}^{B^{\top}} x_{j}^{B,-}+b_{+}^{B}), \quad \xi_{j}^{A} \ge 0 ,\nonumber\\
&\quad \xi_{j}^{B} \ge -f_{j}^{A}(\omega_{+}^{A^{\top}} x_{j}^{A,-}+b_{+}^{A}), \quad \xi_{j}^{B} \ge 0 ,\quad (j\in I^{-})
\end{align}
and
\begin{align}\label{MPWTSVM_Q2}
 \mathbf{\min} \quad
 &\frac{1}{2} \sum_{i=1}^{l_2}\sum_{j=1}^{l_2}{W_{s,ij}^{A}} (\omega_{-}^{A^{\top}} x_{j}^{A,-}+b_{-}^{A})^2
 +\frac{1}{2} \gamma \sum_{i=1}^{l_2}\sum_{j=1}^{l_2}{W_{s,ij}^{B}} (\omega_{-}^{B^{\top}} x_{j}^{B,-}+b_{-}^{B})^2 \nonumber \\
 & +C_{A2}\sum_{i=1}^{l_1}{\xi_{i}^{A}} +C_{B2}\sum_{i=1}^{l_1}{{\xi_{i}^{B}}+C_2\sum_{i=1}^{l_1}{\xi_{i}^{A}\xi_{i}^{B}}} \nonumber \\
 s.t.\quad & f_{i}^{A}(\omega_{-}^{A^{\top}} x_{i}^{A,+}+b_{-}^{A})+\xi_{i}^{A} \ge f_{i}^{A} \cdot 1, \nonumber \\
 & f_{i}^{B}(\omega_{-}^{B^{\top}} x_{i}^{B,+}+b_{-}^{B})+\xi_{i}^{B} \ge f_{i}^{B} \cdot 1, \nonumber \\
 & \xi_{i}^{A} \ge f_{i}^{B}(\omega_{-}^{B^{\top}} x_{i}^{B,+}+b_{-}^{B}), \quad \xi_{i}^{A} \ge 0 , \nonumber \\
 &  \xi_{i}^{B} \ge f_{i}^{A}(\omega_{-}^{A^{\top}} x_{i}^{A,+}+b_{-}^{A}), \quad \xi_{i}^{B} \ge 0 ,\quad (i\in I^{+})
\end{align}
where $C_A,C_B,C,C_{A2},C_{B2},C_2$ are non-negative parameters and $\gamma$ is a non-negative trade-off parameter.

Under the supposition that both perspectives are equally important, the MPWTSVM targets on seeking four hyperplanes, which are explained in detail below:
\begin{enumerate}[(1)]
    \item  The variables to be worked out in problem (\ref{MPWTSVM_Q1}) are $w_+^A, w_+^B, b_+^A,b_+^B, \xi^A$ and $\xi^B$. Problem (\ref{MPWTSVM_Q2}) is similar.
    \item A larger $W_{s,ij}^A$ indicates a larger weight within the positive class of view A, which can make full use of the structural information. The introduction of this term can make a more compact structure and thus have a better generalization performance. View B is the same. These matrices fully exploit the intra-class information of view A and B.
    \item Minimizing $\xi_{i}^{A}, \xi_{i}^{B}$ demonstrates the product of error variables of A and B should be as small as possible.
    Besides, minimizing the coupling term $C \sum_{i=1}^{l}{\xi_{i}^{A}\xi_{i}^{B}}$  indicates that the high amount of error in one perspective can be recompensed by the low amount of error in the other perspective. It means that a bigger error variable can be allowed in one perspective. In this way, the classification results of the models constructed in different views can converge, and the principle of consensus is realized.  
    \item If the inter-class weight $f^A$ or $f^B$ is 0, it means that the sample constraint is redundant and can be deleted, as a consequence the algorithm efficiency is greatly improved.
    \item Our model uses each view separately as privileged information to reformulate the slack variables. By limiting the non-negative slack variables $\xi^A$ and $\xi^B$ through the unknown non-negative correcting functions determined by view A and B, therefore, MPWTSVM realizes the complementary principle.
\end{enumerate}

\subsection{The dual problem}
For simplicity, we let
$d_i^{A/B}=\sum\limits_{i=1}^{l_1} W_{s,ij}^{A/B}$,
$\mathbf{w}_{\pm}^{A/B}=\left(\begin{array}{cc}
     \omega_{\pm}^{A/B} \\
     b_{\pm}
\end{array}\right), 
\mathbf{x}^{A/B}=(x^{A/B},1)$.
Aiming at getting the solution of (\ref{MPWTSVM_Q1}), the corresponding Lagrangian function can be constructed as
\begin{align}\label{L1}
L(\mathbf{w}_{+}^{A}, \mathbf{w}_{+}^{B}, {\xi}_{j}^A,{\xi}_{j}^B)=
& \quad \frac{1}{2}\sum_{i=1}^{l_1}d_{i}^{A}(\mathbf{w}_{+}^{A^T} \mathbf{x}_{i}^{A,+})^2+\frac{1}{2} \gamma \sum_{i=1}^{l_1}d_{i}^{B} (\mathbf{w}_{+}^{B^T} \mathbf{x}_{j}^{B,+})^2 \nonumber\\
& \quad +C_{A}\sum_{j=1}^{l_2}{\xi_{j}^{A}}+C_{B}\sum_{j=1}^{l_2}{{\xi_{j}^{B}}+C\sum_{j=1}^{l_2}{\xi_{j}^{A}\xi_{j}^{B}}}\nonumber\\
&  \quad-\sum_{j=1}^{l_2}\alpha_{j}^{A} \left[-f_{j}^{A} (\mathbf{w}_{+}^{A^T} \mathbf{x}_{j}^{A,-}) + \xi_{j}^{A}-f_{j}^{A}\right]\nonumber\\
&  \quad-\sum_{j=1}^{l_2}\alpha_{j}^{B} \left[-f_{j}^{B} (\mathbf{w}_{+}^{B^T} \mathbf{x}_{j}^{B,-}) + \xi_{j}^{B}-f_{j}^{B}\right]\nonumber\\
&  \quad-\sum_{j=1}^{l_2} \lambda_{j}^{A}\left[\xi_{j}^{A}+f_{j}^{B}(\mathbf{w}_{+}^{B^T} \mathbf{x}_{j}^{B,-})\right] \nonumber\\
& \quad -\sum_{j=1}^{l_2} \lambda_{j}^{B} \left[ \xi_{j}^{B}+f_{j}^{B}(\mathbf{w}_{+}^{A^T} \mathbf{x}_{j}^{A,-}) \right] \nonumber\\
&  \quad-\sum_{j=1}^{l_2}\beta_{j}^{A}\xi_{j}^{A}-\sum_{j=1}^{l_2}\beta_{j}^{B} \xi_{j}^{B},
\end{align}
where $\alpha^A=(\alpha_1^A,...,\alpha_{l_2}^A)^{\top}$,\quad
$\alpha^B=(\alpha_1^B,...,\alpha_{l_2}^B)^{\top}$,\quad
$\lambda^A=(\lambda_1^A,...,\lambda_{l_2}^A)^{\top}$,\quad
$\lambda^B=(\lambda_1^B,...,\lambda_{l_2}^B)^{\top}$,\quad
$\beta^A=(\beta_1^A,...,\beta_{l_2}^A)^{\top}$,\quad
$\beta^B=(\beta_1^B,...,\beta_{l_2}^B)^{\top}$ are the non-negative Lagrange multipliers vectors.

Differentiating the Lagrangian function $L$ with respect to variables $\mathbf{w}_{+}^A, \mathbf{w}_{+}^B$, ${\xi}_j^A, {\xi}_j^B$ yields the following Karush-Kuhn-Tucker (KKT) conditions:
\begin{align}
\label{DUALMPWTSVM_WA}
& \frac{\partial L}{\partial {{\mathbf{w}}_{+}^A}} = \sum_{i=1}^{l_1} d_i^A \mathbf{x}_i^{A,+}  {\mathbf{x}_i^{A,+}}^{\top} \mathbf{w}_{+}^A + \sum_{j=1}^{l_2} \alpha_j^A f_j^A \mathbf{x}_j^{A,-} - \sum_{j=1}^{l_2} \alpha_j^B f_j^A \mathbf{x}_j^{A,-} = 0 ,\\
\label{DUALMPWTSVM_WB}
& \frac{\partial L}{\partial {{\mathbf{w}}_{+}^B}} = \gamma \sum_{i=1}^{l_1} d_i^B \mathbf{x}_i^{B,+}  {\mathbf{x}_i^{B,+}}^{\top} \mathbf{w}_{+}^B + \sum_{j=1}^{l_2} \alpha_j^B f_j^B \mathbf{x}_j^{B,-} - \sum_{j=1}^{l_2} \alpha_j^A f_j^B \mathbf{x}_j^{B,-} = 0 ,\\
\label{beta1}& \frac{\partial L}{\partial {\xi_{j}^A}}=C_A+C\cdot \xi_j^B-\alpha_j^A-\lambda_j^A-\beta_j^A =0,\\
\label{beta2}& \frac{\partial L}{\partial {\xi_{j}^B}}=C_B+C\cdot \xi_j^A-\alpha_j^B-\lambda_j^B-\beta_j^B =0,
\end{align}
\begin{align}
& \alpha_j^A \left(-f_{j}^{A}\mathbf{w}_{+}^{A^{\top}} \mathbf{x_{j}}^{A,-} -f_{j}^{A}+\xi_j^A \right) =0,\\
& \alpha_j^B \left(-f_{j}^{B}\mathbf{w}_{+}^{B^{\top}} \mathbf{x_{j}}^{B,-} -f_{j}^{B}+\xi_j^B \right) =0,\\
& \lambda_j^A \left(\xi_j^A+f_j^B\mathbf{w}_{+}^{B^{\top}} \mathbf{x_{j}}^{B,-} \right) =0,\\
& \lambda_j^B \left(\xi_j^B+f_j^A\mathbf{w}_{+}^{A^{\top}} \mathbf{x_{j}}^{A,-} \right) =0,\\
& \beta_j^A \xi_j^A=0,\\
& \beta_j^B \xi_j^B=0.
\end{align}

Expressing (\ref{DUALMPWTSVM_WA}) and (\ref{DUALMPWTSVM_WB}) in matrix form, we get:
\begin{align}
    & {X_+^A}^{\top} D_+^A X_+^A w_+^A + {X_{-}^A}^{\top} F_{-}^A \alpha_{-}^A - {X_{-}^A}^{\top} F_{-}^A \lambda_{-}^B =0,\\
    & \gamma {X_+^B}^{\top} D_+^B X_+^B w_+^B + {X_{-}^B}^{\top} F_{-}^B \alpha_{-}^B - {X_{-}^B}^{\top} F_{-}^B \lambda_{-}^A =0,
\end{align}
where $D_{+}^A=diag(d_1^A,d_2^A,...,d_{l_1}^A)$,\quad $D_{+}^B=diag(d_1^B,d_2^B,...,d_{l_1}^B)$,\quad \\
$F_{-}^A=diag(f_1^A,f_2^A,...,f_{l_2}^A)$ and
$F_{-}^B=diag(f_1^B,f_2^B,...,f_{l_2}^B)$. \\

Thereupon we can obtain the optimal solutions $w_+^A$ and $w_+^B$ of (\ref{MPWTSVM_Q1}):
\begin{align}
    & w_{+}^A=-{\left({X_+^A}^{\top} D_+^A X_+^A\right)}^{-1}
    \left({X_{-}^A}^{\top} F_{-}^A (\alpha_{-}^A-\lambda_{-}^B )\right),\label{omega+A}\\
    & w_{+}^B=-{\left( \gamma {X_+^B}^{\top} D_+^B X_+^B\right)}^{-1}
    \left({X_{-}^B}^{\top} F_{-}^B (\alpha_{-}^B-\lambda_{-}^A )\right).\label{omega+B}
\end{align}

If the matrix ${X_+^A}^{\top} D_+^A X_+^A$ or $\gamma {X_+^B}^{\top} D_+^B X_+^B$ is irreversible, we can introduce a regularization term $\epsilon I$, $\epsilon >0$, then (\ref{omega+A}) and (\ref{omega+B}) can be written as 
\begin{align}
    & w_{+}^A=-{\left({X_+^A}^{\top} D_+^A X_+^A + \epsilon I\right)}^{-1}
    \left({X_{-}^A}^{\top} F_{-}^A (\alpha_{-}^A-\lambda_{-}^B )\right),\\
    & w_{+}^B=-{\left( \gamma {X_+^B}^{\top} D_+^B X_+^B + \epsilon I\right)}^{-1}
    \left({X_{-}^B}^{\top} F_{-}^B (\alpha_{-}^B-\lambda_{-}^A )\right).
\end{align}
The same method is used if we encounter the problem of matrix integrability in the later part.

 By substituting the above equation into (\ref{L1}), we can derive the dual formulation as follows,
 \begin{align}\label{DMPWTSVM_Q1}
    \mathbf{\max} \quad
    & -\frac{1}{2}
     {\left( \alpha_{-}^A - \lambda_{-}^B \right) }^{\top}
    F_{-}^{A} X_{-}^A {\left({X_+^A}^{\top} D_+^A X_+^A\right)}^{-1}
    {X_{-}^A}^{\top} F_{-}^A (\alpha_{-}^A-\lambda_{-}^B ) \nonumber \\
    & -\frac{1}{2\gamma}{\left(\alpha_{-}^B-\lambda_{-}^A \right)}^{\top}
    F_{-}^{B} X_{-}^B {\left( {X_+^B}^{\top} D_+^B X_+^B\right)}^{-1}
    {X_{-}^B}^{\top} F_{-}^B (\alpha_{-}^B-\lambda_{-}^A ) \nonumber \\
    & +{\alpha_{-}^A}^{\top} F_{-}^A \bm{e}_{-}
    +{\alpha_{-}^B}^{\top} F_{-}^B \bm{e}_{-}
    -C {\xi_{-}^A}^{\top} \xi_{-}^B \nonumber \\
    s.t. \quad
    & \alpha_{-}^A,\alpha_{-}^B,\lambda_{-}^A,\lambda_{-}^B,\beta_{-}^A,\beta_{-}^B \ge 0,
 \end{align}
 where $\xi_{-}^A=\frac{1}{C} \left(\alpha_{-}^B + \lambda_{-}^B + \beta_{-}^B - C_B \cdot \bm{e}_{-}\right)$
 and
    $\xi_{-}^B=\frac{1}{C} \left(\alpha_{-}^A + \lambda_{-}^A + \beta_{-}^A - C_A \cdot \bm{e}_{-}\right)$.

 Due to $\beta_j^A,\beta_j^B \ge 0$ in (\ref{beta1}) and (\ref{beta2}), we have $\alpha_j^A+\lambda_j^A- C \cdot \xi_j^B \le C_A$,
$\alpha_j^B+\lambda_j^B- C \cdot \xi_j^A \le C_B$. Owing to the complexity of objective function   (\ref{DMPWTSVM_Q1}), we can work out the following unanimous dual problem as a substitute for simplification,
\begin{align}\label{DMPWTSVM_QPP1}
    \mathbf{\min} \quad
    & \frac{1}{2}
     {\left( \alpha_{-}^A - \lambda_{-}^B \right) }^{\top}
    F_{-}^{A} X_{-}^A {\left({X_+^A}^{\top} D_+^A X_+^A\right)}^{-1}
    {X_{-}^A}^{\top} F_{-}^A (\alpha_{-}^A-\lambda_{-}^B ) \nonumber \\
    & +\frac{1}{2\gamma}{\left(\alpha_{-}^B-\lambda_{-}^A \right)}^{\top}
    F_{-}^{B} X_{-}^B {\left( {X_+^B}^{\top} D_+^B X_+^B\right)}^{-1}
    {X_{-}^B}^{\top} F_{-}^B (\alpha_{-}^B-\lambda_{-}^A ) \nonumber \\
    & -{\alpha_{-}^A}^{\top} F_{-}^A \bm{e}_{-}
    -{\alpha_{-}^B}^{\top} F_{-}^B \bm{e}_{-}
    +C {\xi_{-}^A}^{\top} \xi_{-}^B \nonumber \\
    s.t. \quad
    & \alpha_-^A+\lambda_-^A- C \cdot \xi_-^B \le C_A \cdot \bm{e}_{-} \nonumber\\
    & \alpha_-^B+\lambda_-^B- C \cdot \xi_-^A \le C_B \cdot \bm{e}_{-} \nonumber\\
    & \alpha_{-}^A,\alpha_{-}^B,\lambda_{-}^A,\lambda_{-}^B,\beta_{-}^A,\beta_{-}^B \ge 0\cdot\bm{e}_{-}.
\end{align}

Next, we define $\pi_+ = ({\alpha_{-}^A}^{\top}, {\alpha_{-}^B}^{\top}, {\lambda_{-}^A}^{\top}, {\lambda_{-}^B}^{\top} , {\xi_{-}^A}^{\top}, {\xi_{-}^B}^{\top})^{\top}$.
Concisely, (\ref{DMPWTSVM_QPP1}) can be further reformulated as

\begin{equation}
\begin{aligned}\label{MPWTSVM-1}
&\underset{\pi^{+}}{\text{min}} \quad \frac{1}{2} \pi_{+}^{\top} H_{+} \pi_{+} + p_{+}^{\top} \pi_{+}  \\
& s.t. \quad A_+\pi_+ \le b_+, \quad \pi_{+} \ge 0,
 \end{aligned}
\end{equation}
where \\
\begin{align}
& H_{+}=
\begin{pmatrix}
    H_1^+ & 0_{l_2} & 0_{l_2} & -H_1^+ & 0_{l_2} & 0_{l_2}\\\\
    0_{l_2} & H_2^+ & -H_2^+ & 0_{l_2} & 0_{l_2} & 0_{l_2}\\\\
    0_{l_2} & -H_2^+ & H_2^+ & 0_{l_2} & 0_{l_2} & 0_{l_2}\\\\
    -H_1^+ & 0_{l_2} & 0_{l_2} & H_1^+ & 0_{l_2} & 0_{l_2}\\\\
    0_{l_2} & 0_{l_2} & 0_{l_2} & 0_{l_2} & 0_{l_2} & C\cdot E_{l_2}\\\\
    0_{l_2} & 0_{l_2} & 0_{l_2} & 0_{l_2} & C\cdot E_{l_2} & 0_{l_2}
\end{pmatrix}_{6l_2 \times 6l_2}, \nonumber\\
& H_1^+ = \left(F_{-}^{A} X_{-}^A {\left({X_+^A}^T D_+^A X_+^A\right)}^{-1}
{X_{-}^A}^T F_{-}^A\right),\nonumber\\
& H_2^+ = \left(\frac{1}{\gamma} F_{-}^{B} X_{-}^B {\left( {X_+^B}^T D_+^B X_+^B\right)}^{-1}{X_{-}^B}^T F_{-}^B\right), \nonumber\\
& p_+^T = \left(-e_{-}^T F_{-}^A \quad -e_{-}^T F_{-}^B \quad 0_{1\times l_2}\quad 0_{1\times l_2} \quad 0_{1\times l_2} \right)_{1\times 6l_2}, \nonumber
\end{align}
\begin{align}
& A_+ =
\begin{pmatrix}
    E_{l_2} & 0_{l_2} & E_{l_2} & 0_{l_2} & 0_{l_2} & - C\cdot E_{l_2}\\\\
    0_{l_2} & E_{l_2} & 0_{l_2} & E_{l_2} & -C\cdot E_{l_2} & 0_{l_2}
\end{pmatrix}_{2l_2\times 6l_2},\nonumber\\
& b_+^T = \left( C_A \cdot e_{-}^T \quad C_B \cdot e_{-}^T\right)_{1\times 2l_2},\nonumber
\end{align}
$E_{l_2}$ is the $l_2\times l_2$ identity matrix, $0_{l_2}$ is the $l_2 \times l_2$ matrix with all entries be 0 and $e_{-}$ is a column vector with the proper dimension with element 1.

Using a similar process, the second optimization problem (\ref{MPWTSVM_Q2}) can be written as:

\begin{equation}
\begin{aligned}\label{MPWTSVM-2}
&\underset{\pi^{-}}{\text{min}} \quad \frac{1}{2} \pi_{-}^{\top} H_{-} \pi_{-} + p_{-}^{\top} \pi_{-}  \\
& s.t. \quad A_-\pi_- \le b_-, \quad \pi_{-} \ge 0,
 \end{aligned}
\end{equation}
where $\pi_- = ({\alpha_{+}^A}^{\top},  {\alpha_{+}^B}^{\top}, {\lambda_{+}^A}^{\top}, {\lambda_{+}^B}^{\top}, {\xi_{+}^A}^{\top}, {\xi_{+}^B}^{\top})^{\top}$,
\begin{align}
H_{-}=
\begin{pmatrix}
    H_1^- & 0_{l_1} & 0_{l_1} & -H_1^- & 0_{l_1} & 0_{l_1}\\\\
    0_{l_1} & H_2^- & -H_2^- & 0_{l_1} & 0_{l_1} & 0_{l_1}\\\\
    0_{l_1} & -H_2^- & H_2^- & 0_{l_1} & 0_{l_1} & 0_{l_1}\\\\
    -H_1^- & 0_{l_1} & 0_{l_1} & H_1^- & 0_{l_1} & 0_{l_1}\\\\
    0_{l_1} & 0_{l_1} & 0_{l_1} & 0_{l_1} & 0_{l_1} & C_2\cdot E_{l_1}\\\\
    0_{l_1} & 0_{l_1} & 0_{l_1} & 0_{l_1} & C_2\cdot E_{l_1} & 0_{l_1}
\end{pmatrix}_{6l_1 \times 6l_1}, \nonumber
\end{align}
\begin{align}
& H_1^- = \left(F_{+}^{A} X_{+}^A {\left({X_-^A}^{\top} D_-^A X_-^A\right)}^{-1}
{X_{+}^A}^{\top} F_{+}^A\right),\nonumber\\
& H_2^- = \left(\frac{1}{\gamma_2} F_{+}^{B} X_{+}^B {\left( {X_-^B}^{\top} D_-^B X_-^B\right)}^{-1}{X_{+}^B}^{\top} F_{+}^B\right), \nonumber\\
& p_-^{\top} = \left(-e_{+}^{\top} F_{+}^A \quad -e_{+}^{\top} F_{+}^B \quad 0_{1\times l_1}\quad 0_{1\times l_1} \quad 0_{1\times l_1} \right)_{1\times 6l_1}, \nonumber \\
& A_- =
\begin{pmatrix}
    E_{l_1} & 0_{l_1} & E_{l_1} & 0_{l_1} & 0_{l_1} & - C_2\cdot E_{l_1}\\\\
    0_{l_1} & E_{l_1} & 0_{l_1} & E_{l_1} & -C_2\cdot E_{l_1} & 0_{l_1}
\end{pmatrix}_{2l_1\times 6l_1},\nonumber\\
& b_-^{\top} = \left( C_{A2} \cdot e_{+}^{\top} \quad C_{B2} \cdot e_{+}^{\top}\right)_{1\times 2l_1},\nonumber
\end{align}
$E_{l_1}$ is the $l_1\times l_1$ identity matrix, $0_{l_1}$ is the $l_1 \times l_1$ matrix with all entries be 0 and $e_{+}$ is a column vector with the proper dimension with element 1.

A new data point $x\in \mathbb{R}^n$ is assigned to class r (r=1,2), depending on which plane it is nearer to.
We first have the decision function of view A and view B respectively:
\begin{equation}\label{classA}
class^A(x_A)=\mathop{arg}_{r=1,2}  min\left(\frac{|{x_A}^{\top} w_r^A+b_r^A|}{||w_r^A||}(x_A)\right)
\end{equation}
and
\begin{equation}\label{classB}
    class^B(x_B)=\mathop{arg}_{r=1,2}  min\left(\frac{|{x_B}^{\top} w_r^B+b_r^B|}{||w_r^B||}(x_B)\right).
\end{equation}

Then the decision function combining two views can be given below:
\begin{equation}\label{class}
class(x)=\mathop{arg}_{r=1,2}  min\left(d_r(x)\right),
\end{equation}
where $$d_r(x)=\frac{1}{2}\left( \frac{|{x_A}^{\top} w_r^A+b_r^A|}{||w_r^A||}+\frac{|{x_B}^{\top} w_r^B+b_r^B|}{||w_r^B||} \right).$$

For the sake of perspicuity, we explicitly express the MPWTSVM algorithm in Algorithm \ref{alg1}.
\renewcommand{\algorithmicrequire}{\textbf{Input:}}
\renewcommand{\algorithmicensure}{\textbf{Output:}}
\begin{algorithm}
\caption{QP Algorithm for MPWTSVM.}
\label{alg1}
\begin{algorithmic}[1]
\REQUIRE 
Training datasets $S=\{(\mathbf{x}_i^A,\mathbf{x}_i^B,y_i)\}_{i=1}^l=\{((x_i^A;1),(x_i^B;1),y_i)\}_{i=1}^l$, where label $y_i\in\{-1,1\}$ and the testing sample, $\mathbf{x}$;\\
Initial parameters $\gamma, C_A, C_B, C, D \ge 0$.
\ENSURE 
Decision function as in (\ref{classA}),(\ref{classB}),and(\ref{class}).\\
1. Choose two appropriate kernels $\mathcal{K}_A(x_i^A,x_j^A)$, 
$\mathcal{K}_B(x_i^B,x_j^B)$ and initialize the kernel parameters.

2. Establish and solve QPPs of (\ref{MPWTSVM-1}) and (\ref{MPWTSVM-2}) by using 5-fold cross-validation and choose the best parameters.

3. Construct the separating hyperplanes ${\mathbf{w}_+^A}^{\top} \phi_A (x_A) = 0$,\\ ${\mathbf{w}_-^A}^{\top} \phi_A (x_A) = 0$, ${\mathbf{w}_{+}^B}^{\top} \phi_B (x_B) = 0$,
${\mathbf{w}_{-}^B}^{\top} \phi_B (x_B) = 0$, and\\
 $0.5({\mathbf{w}_+^A}^{\top} \phi_A (x_A)+{\mathbf{w}_-^A}^{\top} \phi_A (x_A)) + 0.5({\mathbf{w}_+^B}^{\top} \phi_A (x_B)+{\mathbf{w}_-^B}^{\top} \phi_A (x_B))= 0$.

4. For a new testing point $\mathbf{x}$, predict its label according to the decision functions (\ref{classA}) or (\ref{classB}) respectively and (\ref{class}) collectively.
\end{algorithmic}
\end{algorithm}

\subsection{Nonlinear MPWTSVM}
In this section, we extend linear MPWTSVM to the nonlinear case.
The kernel-generated hyperplanes are:
\begin{align}
&\mathcal{K}(x_+^A,C^A)w_+^A+b_+^A=0; \quad
\mathcal{K}(x_-^A,C^A)w_-^A+b_-^A=0;\\
&\mathcal{K}(x_+^B,C^B)w_+^B+b_+^B=0; \quad
\mathcal{K}(x_-^B,C^B)w_-^B+b_-^B=0;
\end{align}
where $\mathcal{K}$ is a chosen kernel function defined by $\mathcal{K}(x_i, x_j) = \left(\phi(x_i) \cdot \phi(x_j)\right)$.
$\phi(\cdot)$ is a nonlinear mapping that maps the low-dimensional feature space to the high-dimensional feature space in a non-linear manner.
C denotes training examples from view A and view B respectively.
$C^A=[X_1^A;X_2^A]$ and $C^B=[X_1^B;X_2^B]$, so that positive examples from view t\ (t=A,B) are denoted as $\Omega_+^t$ and negative examples from view t are denoted as $\Omega_-^t$.

We can define:
$$\Omega_+^t=\mathcal{K}(x_+^t,C^t),\quad
\Omega_-^t=\mathcal{K}(x_-^t,C^t),$$
$$\mathbf{w_+^t}=
\left(\begin{array}{l}
\omega_+^t \\
 b_+^t
\end{array} \right),\quad
\mathbf{w_-^t}=
\left(\begin{array}{l}
\omega_-^t \\
 b_-^t
\end{array} \right).$$

In order to simplify the calculation, we update the matrices above:
$$\Omega_+^t=\left(\mathcal{K}(x_+^t,C^t),e_+\right);
\quad \Omega_-^t=\left(\mathcal{K}(x_-^t,C^t),e_-\right).$$

Then the optimization problems for non-linear MPWTSVM can be
formulated as

\begin{align}
\mathbf{\min} \quad
 &\frac{1}{2} \sum_{i=1}^{M}\sum_{j=1}^{M}{W_{s,ij}^{A_1}} (\mathbf{w}_+^{{A}^{\top}} \Omega_i^{A,+})^2
 +\frac{1}{2} \gamma \sum_{i=1}^{M}\sum_{j=1}^{M}{W_{s,ij}^{B_1}} (\mathbf{w}_+^{{B}^{\top}} \Omega_i^{B,+})^2  \nonumber\\
 & +C_{A}\sum_{j=1}^{N}{\xi_{j}^{A}} +C_{B}\sum_{j=1}^{N}{{\xi_{j}^{B}}+C\sum_{j=1}^{N}{\xi_{j}^{A}\xi_{j}^{B}}}\nonumber\\
s.t.
\quad & -f_{j}^{A}(\mathbf{w}_+^{{A}^{\top}} \Omega_j^{A,-})+\xi_{j}^{A} \ge f_{j}^{A} \cdot 1, \nonumber\\
& -f_{j}^{B}(\mathbf{w}_+^{{B}^{\top}} \Omega_j^{B,-})+\xi_{j}^{B} \ge f_{j}^{B} \cdot 1, \nonumber\\
&\quad \xi_{j}^{A} \ge -f_{j}^{B}(\mathbf{w}_+^{{B}^{\top}} \Omega_j^{B,-}), \quad \xi_{j}^{A} \ge 0 ,\nonumber\\
&\quad \xi_{j}^{B} \ge -f_{j}^{A}(\mathbf{w}_+^{{A}^{\top}} \Omega_j^{A,-}), \quad \xi_{j}^{B} \ge 0 ,\quad (j\in I^{-});
\end{align}
and
\begin{align}
 \mathbf{\min} \quad
 &\frac{1}{2} \sum_{i=1}^{N}\sum_{j=1}^{N}{W_{s,ij}^{A_2}} (\mathbf{w}_-^{{A}^{\top}} \Omega_j^{A,-})^2
 +\frac{1}{2} \gamma \sum_{i=1}^{N}\sum_{j=1}^{N}{W_{s,ij}^{B_2}} (\mathbf{w}_-^{{B}^{\top}} \Omega_j^{B,-})^2 \nonumber \\
 & +C_{A2}\sum_{i=1}^{M}{\xi_{i}^{A}} +C_{B2}\sum_{i=1}^{M}{{\xi_{i}^{B}}+C_2\sum_{i=1}^{M}{\xi_{i}^{A}\xi_{i}^{B}}} \nonumber \\
 s.t.\quad & f_{i}^{A}(\mathbf{w}_-^{{A}^{\top}} \Omega_i^{A,+})+\xi_{i}^{A} \ge f_{i}^{A} \cdot 1, \nonumber \\
 & f_{i}^{B}(\mathbf{w}_-^{{B}^{\top}} \Omega_i^{B,+})+\xi_{i}^{B} \ge f_{i}^{B} \cdot 1, \nonumber \\
 & \xi_{i}^{A} \ge f_{i}^{B}(\mathbf{w}_-^{{B}^{\top}} \Omega_i^{B,+}), \quad \xi_{i}^{A} \ge 0 , \nonumber \\
 &  \xi_{i}^{B} \ge f_{i}^{A}(\mathbf{w}_-^{{A}^{\top}} \Omega_i^{A,+}), \quad \xi_{i}^{B} \ge 0 ,\quad (i\in I^{+}).
\end{align}

\section{Comparison with other algorithms}
In this section, we compare our MPWTSVM with SVM-2K \cite{SVM-2K2006}, MVTSVM \cite{MVTSVM2015}, MCPK \cite{MCPK2018} and PSVM-2V \cite{PSVM-2V2017}. The complexity analysis is also included in the following comparisons.
For simplicity, we suppose the number of samples of each class are equal, namely $l_1=l_2=l/2$, where $l$ represents the number of training samples. 
Problem (\ref{MPWTSVM-1}) and (\ref{MPWTSVM-2}) involve two convex QPPs. Both of them can be worked out by the classical QP solver with a time complexity less than $2\mathcal{O}((3l)^3)$ for the reason that the inter-class weight matrix can remove redundant samples and greatly reduce the time complexity.

\subsection{MPWTSVM vs. SVM-2K}
SVM-2K solves a QPP, which combines two-stage learning and SVM into a single optimization. Moreover, it only satisfies the consistency principle of multi-view training by using Kernel Canonical Correlation Analysis (KCCA) theory \cite{CCA} and does not satisfy the principle of complementarity. 
The time complexity of SVM-2K is $\mathcal{O}((4l)^3)$. Our model has better efficiency compared to SVM-2K. 
Our MPWTSVM solves two QPPs which makes it work faster than SVM-2K. Besides, the introduction of KNN makes it more efficient to identify the potential support vector. Based on satisfying the principle of consistency, we met the principle of complementarity with the help of privileged information.

\subsection{MPWTSVM vs. MVTSVM}
Similar to MVTSVM, our model solves two QPPs. The time complexity of MVTSVM is about $2\times \mathcal{O}((2l)^3)$. MVTSVM combines two views by bringing in the similarity constraint between the two-dimensional projections of two different TSVMs from the two feature spaces. It is only applied to the classification of the two views, which cannot solve the general multi-view problem. The supplementary information between different views cannot be effectively used so that the model does not satisfy the principle of complementarity. 
Our model makes full use of each perspective as privileged information to redefine slack variables, thereby satisfying the principle of complementarity.

\subsection{MPWTSVM vs. MCPK}
Compared with MCPK, our model shares with it that they both satisfy both consistency and complementarity principles. The difference is that MCPK solves one QPP, while MPWTSVM solves two problems. The time complexity of MCPK is $\mathcal{O}((6l)^3)$. Our model has better efficiency compared to MCPK. In addition, our model extends the weighting idea of WLTSVM to different perspectives to measure as much similarity information between samples as possible and obtain higher accuracy.

\subsection{MPWTSVM vs. PSVM-2V}
Both PSVM-2V and our model satisfy the complementarity and the consistency principle simultaneously.  PSVM-2V solves a QPP. The time complexity of PSVM-2V is $\mathcal{O}((6l)^3)$. Our model has better efficiency compared to PSVM-2V. PSVM-2V uses regularization terms to limit the differences of prediction results from different perspectives to achieve the consistency principle, which is achieved by the coupling terms in the objective function of our model.
 PSVM-2V and our model realize the principle of complementarity both with the help of privileged information. However, compared to PSVM-2V, our model also draws on the idea of weighting, which enables better preservation of the inter and intra connections and differences of different views in the data.

\section{Experiments}

In this section, we make comparisons between our MPWTSVM and five benchmark methods, SVM+, SVM-2K, MVTSVM, PSVM-2V, and MCPK. We verify the performance of MPWTSVM for binary classification on 45 datasets obtained from \emph{Animals with Attributes} (\emph{AwA})\footnote{Available at https://cvml.ist.ac.at/AwA2/.}\cite{AwA}.
In order to eliminate the influence of size and simplify the numerical calculation, we scale all the features to the range of [0, 1] in the data preprocessing. The experiments are conducted in Matlab R2015a on Windows 7 running on a PC with system configuration Inter(R) Core(TM) i7-6700 CPU (3.40GHz) with 8.00 GB of RAM.

\subsection{Experimental setup}
\textbf{Dataset.} The AwA dataset consists of 30,475 images in 50 animal categories, each image has six pre-extracted features re-represented.
The detailed characteristics of the data set are demonstrated in Table \ref{dataAwA}. Similar to the study\cite{Awaother}, we use the ten classes in \emph{AwA} dataset, i.e. \emph{antelope, grizzly bear, killer whale, beaver, dalmatian, Persian cat, horse, German shepherd, blue whale and Siamese cat}.
Through the one-to-one strategy, we randomly select 200 samples in each class for training and train 45 binary classifiers for each class pair combination.
\begin{table}[h]
{\footnotesize%
\caption{Comprehensive information of datasets we use in the experiments.}
\label{dataAwA}
\begin{tabular}{cccccc}
\hline
Data set & \#Data & \#Classes &
\begin{tabular}[c]{@{}c@{}}
\#Features\\ (View A)
\end{tabular}&
\begin{tabular}[c]{@{}c@{}}
\#Features\\(View B)
\end{tabular} & \#Binary data sets\\ \hline
AwA & 6249 & 10 & 252 & 2000 & \begin{tabular}[c]{@{}c@{}}45\\ (one-versus-one)\end{tabular} \\ \hline
\end{tabular}}
\end{table}\\
\textbf{Kernels.} In these experiments, we choose the
Gaussian radial basis function (RBF)  $\mathcal{K}\left(x_i, x_j\right)= exp(-||x_i-x_j||^2/ \sigma^2)$ for all the algorithms since the RBF is most widely used in the classification problem \cite{MVLSSVM2018,pinball2018,MCPK2018,Subspace2017,Multi-View_Kernel_Spectral_Clustering2018}.\\
\textbf{Measures.} We assess the performance of the classifier by the test accuracy. We perform the grid search strategy and 5-fold cross-validation for all datasets to select the optimal parameters.
For methods other than SVM+, we ponder on the mixed prediction function $sign(0.5(f_A(x_A)+f_B(x_B)))$ in addition to two views' prediction functions $sign(f_A(x_A))$ and $sign(f_B(x_B))$, and choose the one who has the highest precision.\\
\textbf{Benchmark method.}
We make comparisons between the proposed method and five of the most recent methods:
\begin{enumerate}[1)]
\item{${\text{SVM}}_{+}:$}
The SVM+ algorithm \cite{svm+2009} replaces the standard SVM's slack variable by using the non-negative correction function determined by the privilege information. During the training process, we separately use the two views as privileged information for each other.
\item{SVM-2K:}
SVM-2K combines two-stage learning---KCCA followed by SVM, into a single optimization \cite{SVM-2K2006}.
\item{MVTSVM:}
MVTSVM merges two perspectives by introducing similarity constraints between two one-dimensional projections \cite{MVTSVM2015}. The model learns two hyperplanes and solves a pair of QPPs rather than one.
\item{PSVM-2V:}
PSVM-2V extends LUPI (learning using privileged information) to MVL \cite{PSVM-2V2017}.
\item{MCPK:}
MCPK is a simple and effective approach to MVL coupling privilege kernels and satisfies two principles for MVL \cite{MCPK2018}.\\
\end{enumerate}
\textbf{Parameters.}
For all algorithms, the optimal parameters are decided by ﬁve-fold cross validation. The parameter C in $\text{SVM}_+$ is selected from the set $\{ 10^{-3}, 10^{-2}, 10^{-1}, 1, 10^1, 10^2, 10^3\}$. Penalty parameters $C_A,C_B$ and $C$ of SVM-2K and PSVM-2V are selected from $\{ 10^{-3}, 10^{-2}, 10^{-1}, 1, 10^1, 10^2, 10^3\}$.
 For MVTSVM, we let $C_1=C_2=C_3=C_4$ and $D=H$, and both of them are turned over the set $\{ 10^{-3}, 10^{-2}, 10^{-1}, 1, 10^1, 10^2, 10^3\}$.
 For MCPK, we set $C_A=C_B=D$ varying in the set $\{ 10^{-3}, 10^{-2}, 10^{-1}, 1, 10^1, 10^2, 10^3\}$.
 For MPWTSVM, we set $C_A=C_B=C=D$ from $\{ 10^{-3}, 10^{-2}, 10^{-1}, 1, 10^1, 10^2, 10^3\}$. The neighborhood size k is searched within $\{3,5,7,9,11\}$.
Moreover, the trade-off parameter $\gamma$ in $\text{SVM}_+$, PSVM-2V, MCPK and MPWTSVM is chosen from the set $\{ 10^{-3}, 10^{-2}, 10^{-1}, 1,10^1, 10^2, 10^3\}$.
 The kernel parameter $\sigma$ for RBF kernel function is chosen from $\{ 10^{-3}, 10^{-2}, 10^{-1}, 1, 10^1, 10^2, 10^3\}$.
 For the sake of simplicity, in the multi-view model, the kernel parameters of the two views are set to be the same.

\subsection{Parameter analysis}
To study the influence of parameters, we examine the parameter sensitivity of MPWTSVM on nine datasets of AwA. The ranges of variation parameters C, $\gamma$, $\sigma$ are the same as that given in Section 5.1, and we calculate the test data's accuracy. 

\begin{figure}[h]
\centering
\subfigure[gb. vs bea.]{
\begin{minipage}[t]{0.34\linewidth}
\centering
\includegraphics[width=1.5in,trim=100 250 50 275,clip]{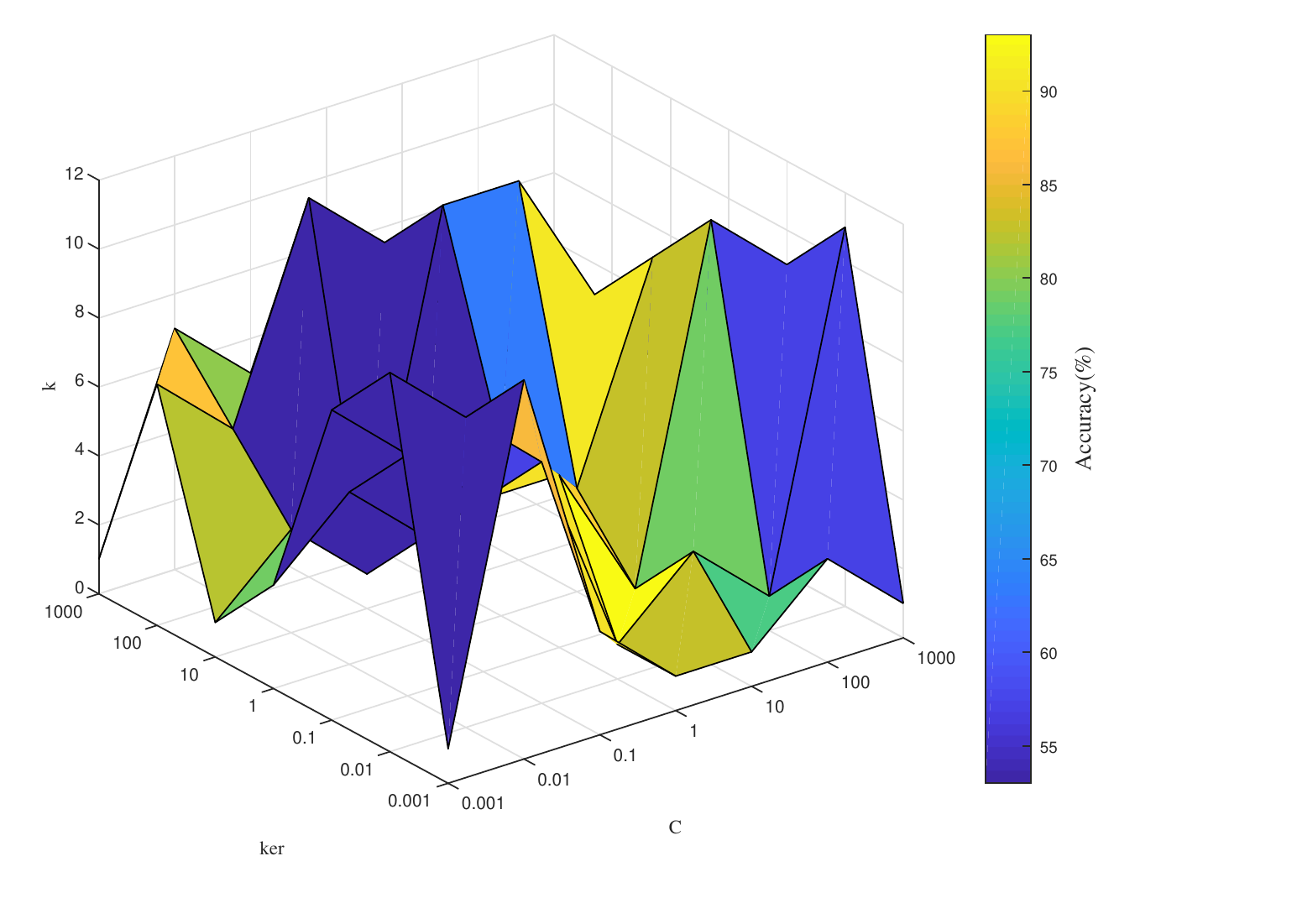}
\end{minipage}%
}%
\subfigure[gb. vs dal.]{
\begin{minipage}[t]{0.34\linewidth}
\centering
\includegraphics[width=1.5in,trim=100 250 50 275,clip]{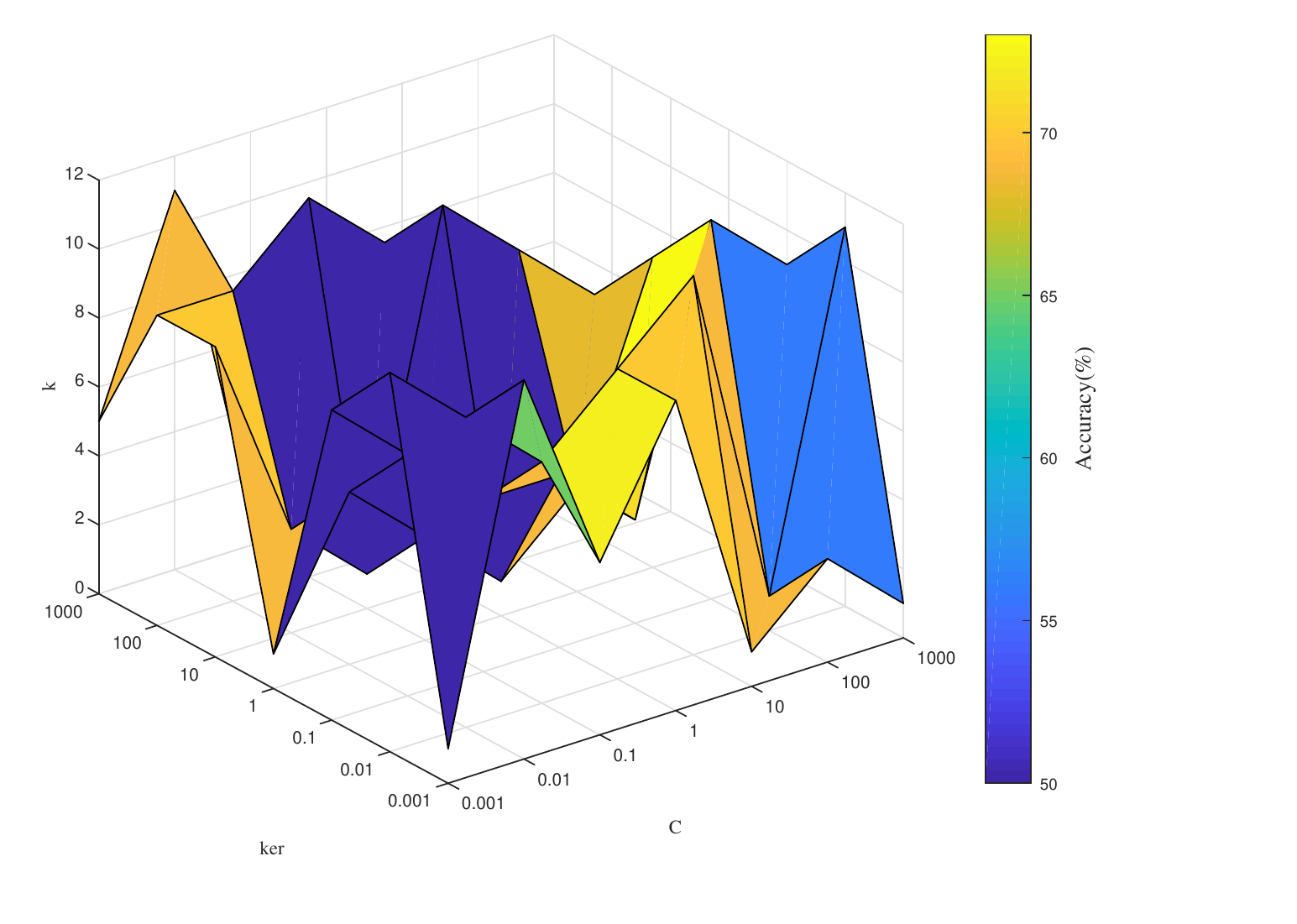}
\end{minipage}%
}%
\subfigure[bea. vs sc.]{
\begin{minipage}[t]{0.34\linewidth}
\centering
\includegraphics[width=1.5in,trim=100 250 50 275,clip]{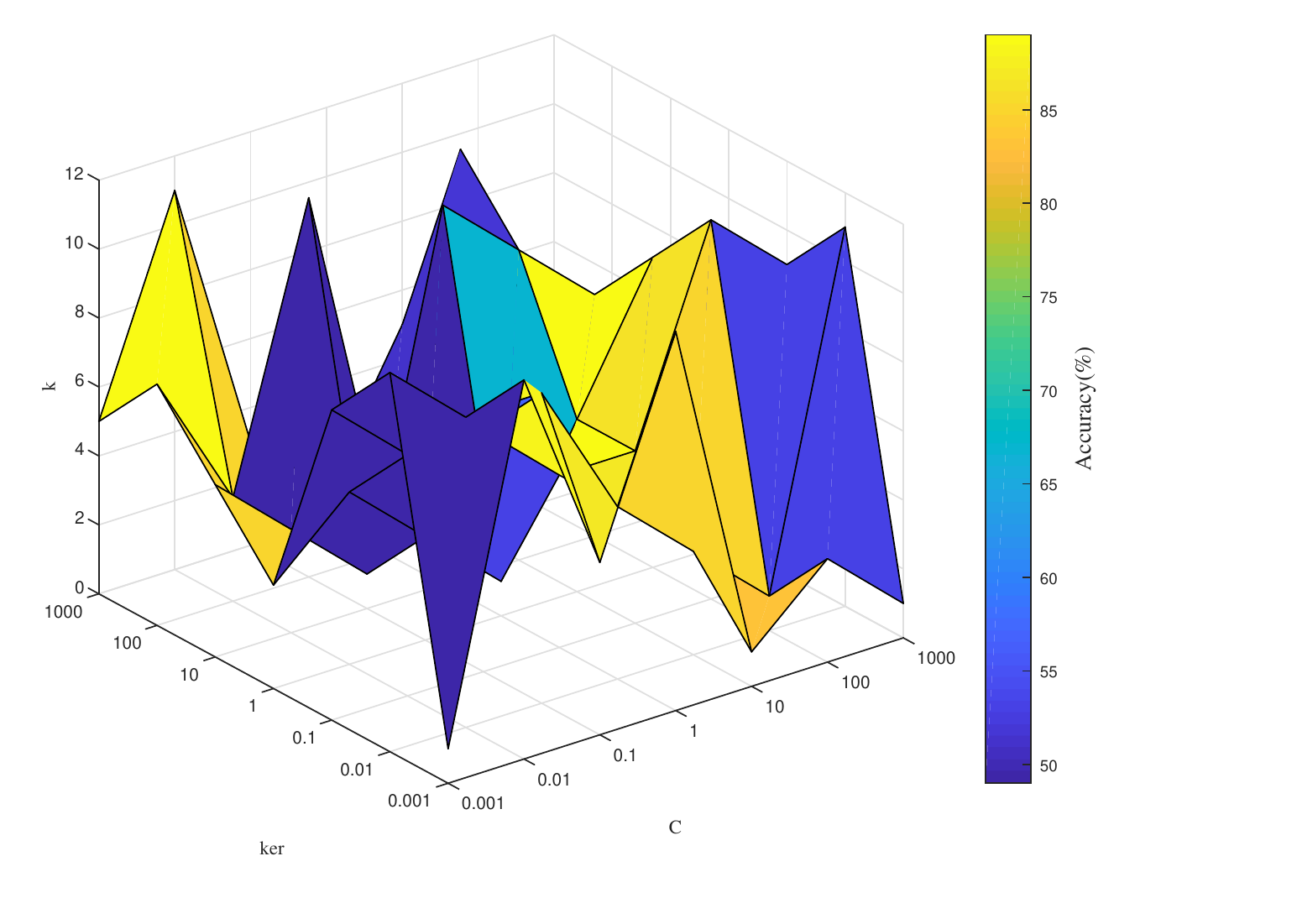}
\end{minipage}
}%

\subfigure[bea. vs dal.]{
\begin{minipage}[t]{0.34\linewidth}
\centering
\includegraphics[width=1.5in,trim=100 250 50 275,clip]{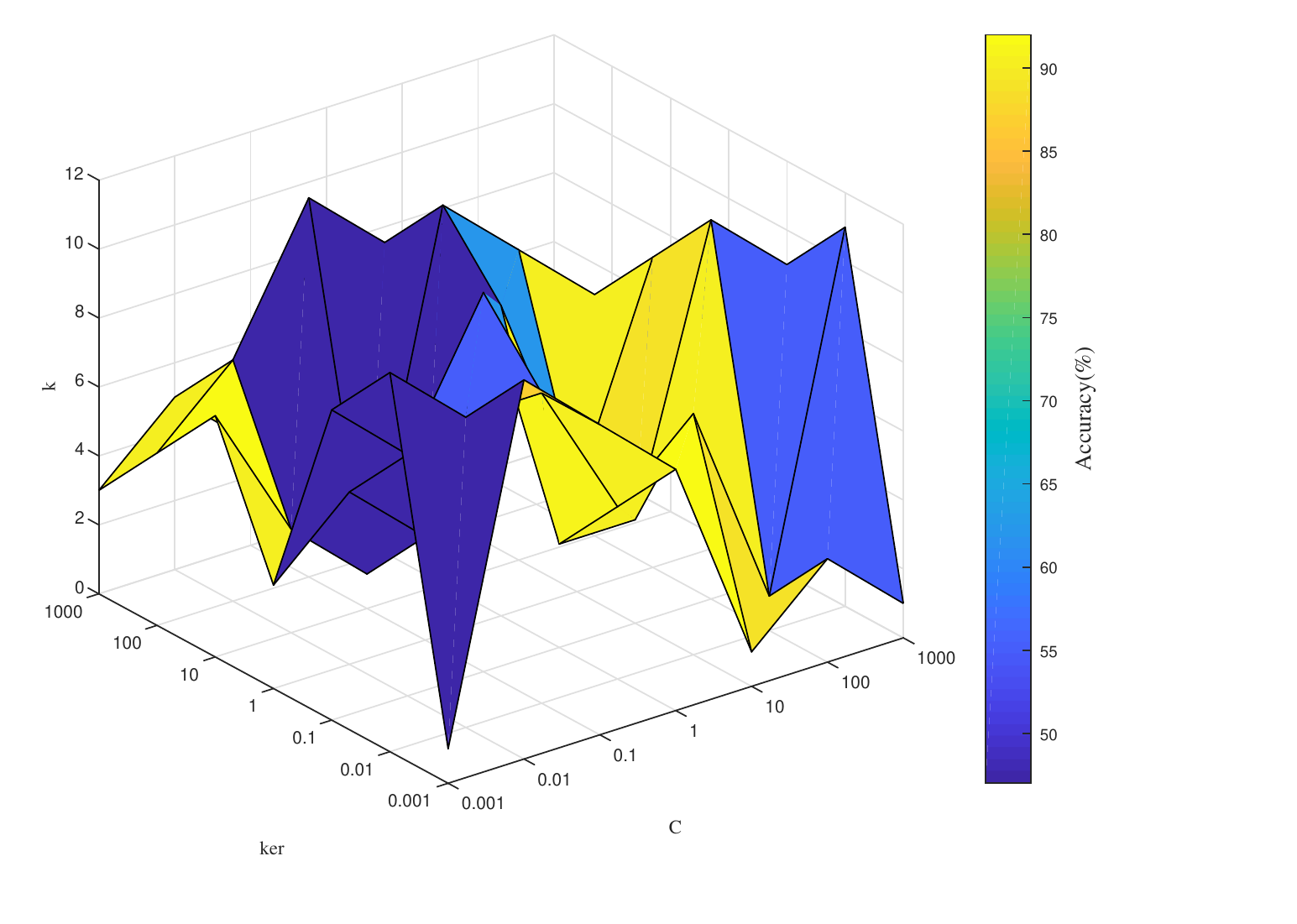}
\end{minipage}
}%
\subfigure[bea. vs pt.]{
\begin{minipage}[t]{0.34\linewidth}
\centering
\includegraphics[width=1.5in,trim=100 250 50 275,clip]{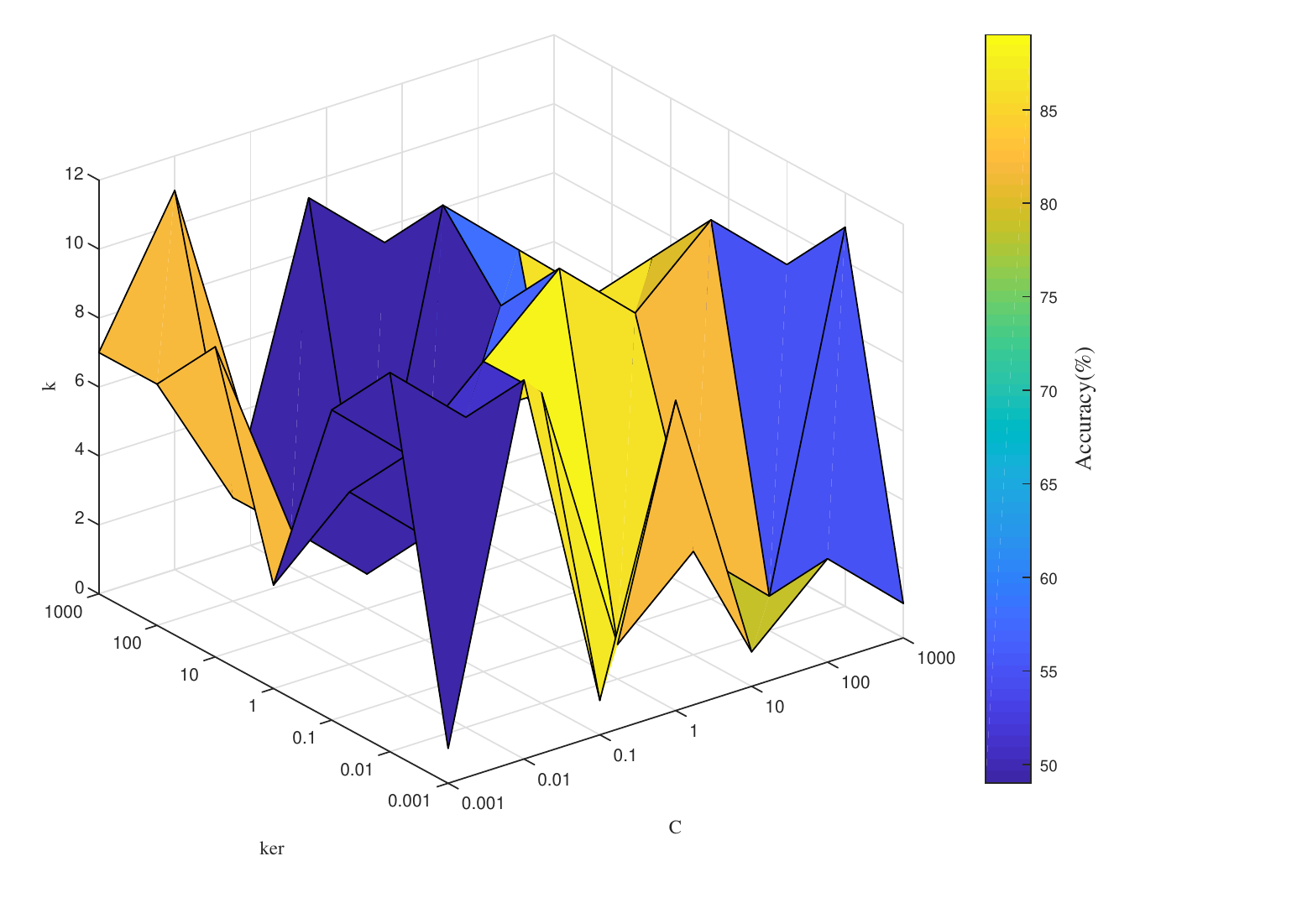}
\end{minipage}%
}%
\subfigure[bea. vs ho.]{
\begin{minipage}[t]{0.34\linewidth}
\centering
\includegraphics[width=1.5in,trim=100 250 50 275,clip]{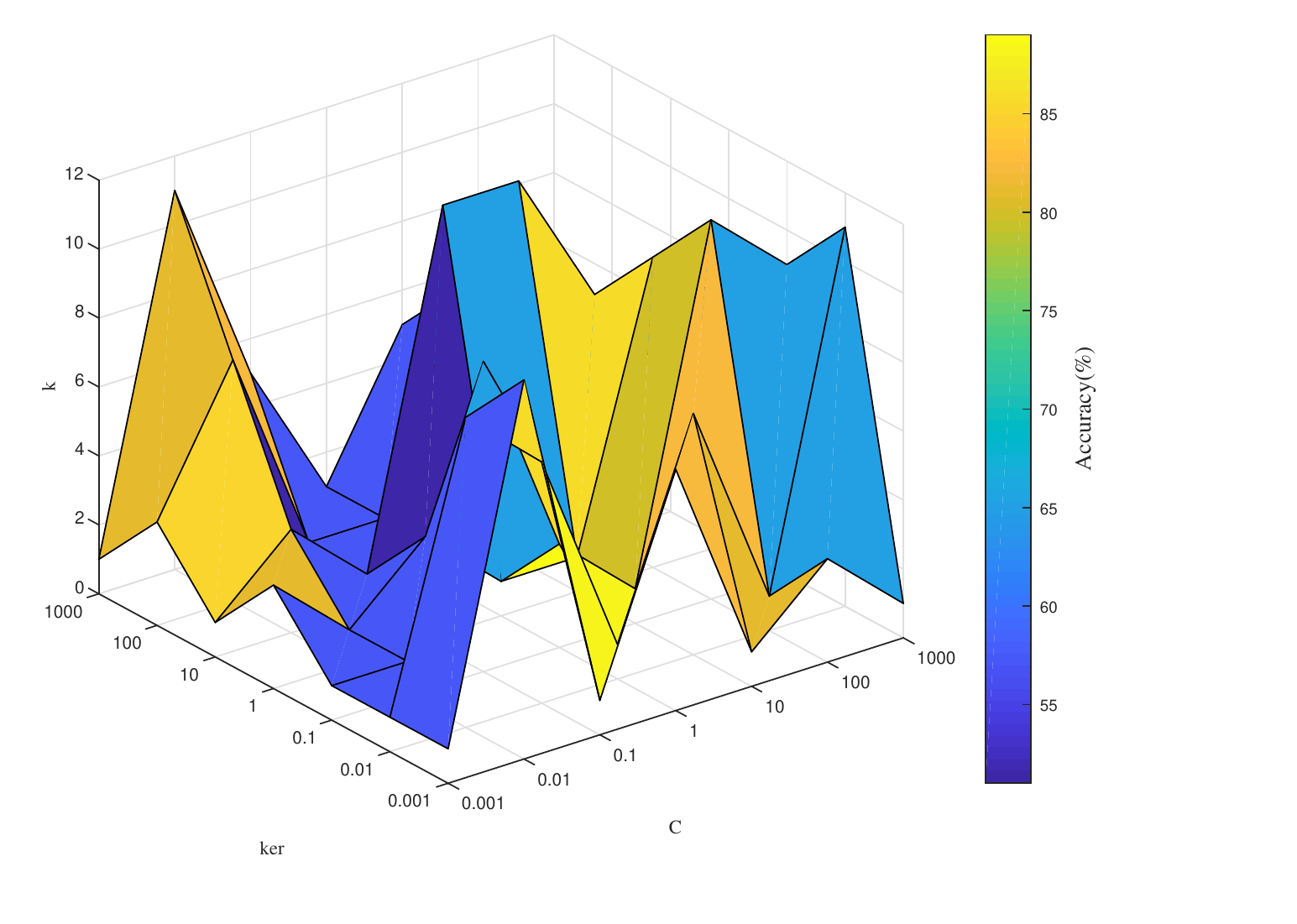}
\end{minipage}
}

\subfigure[bea. vs Gs.]{
\begin{minipage}[t]{0.34\linewidth}
\centering
\includegraphics[width=1.5in,trim=100 250 50 275,clip]{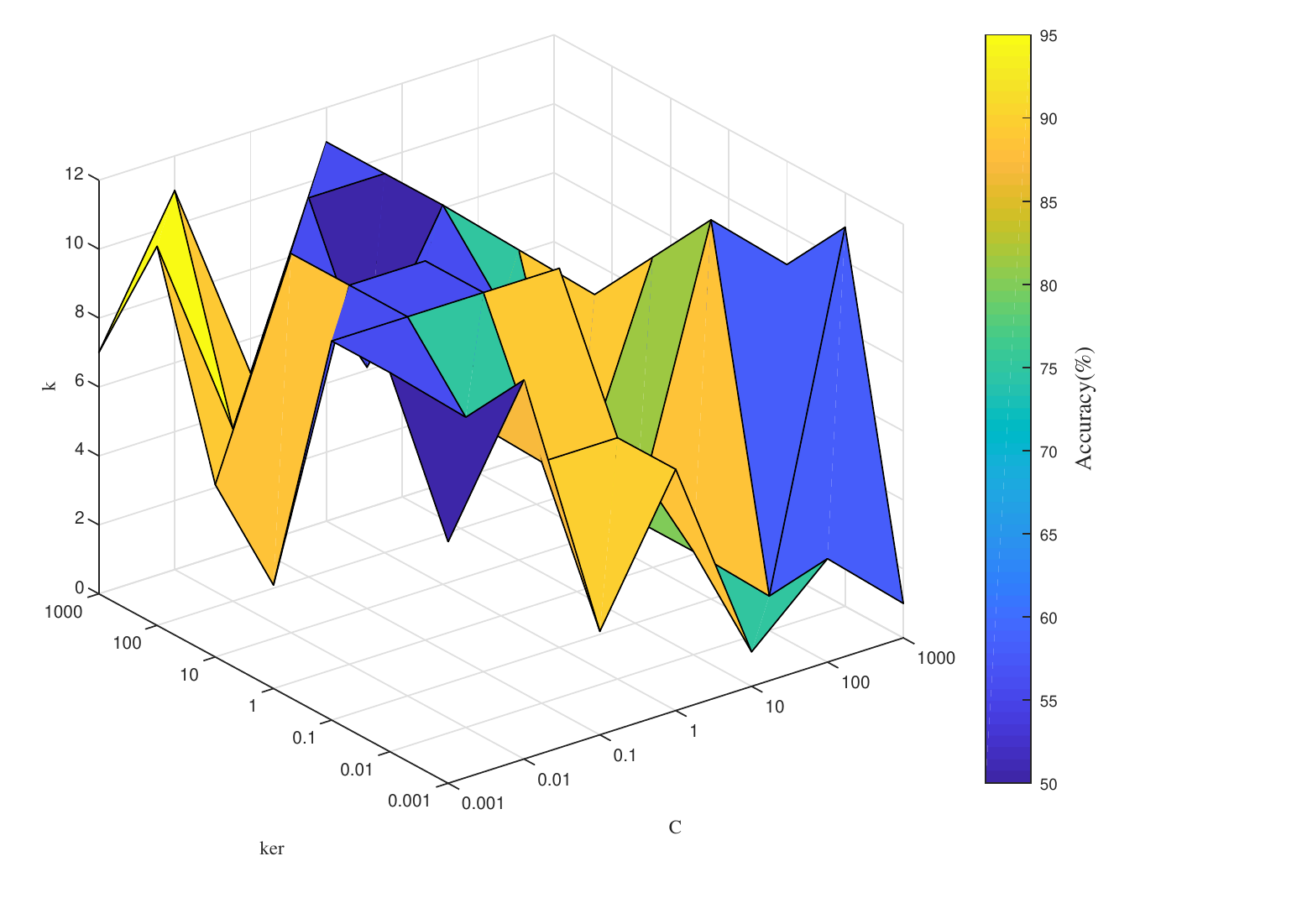}
\end{minipage}%
}%
\subfigure[bea. vs bw.]{
\begin{minipage}[t]{0.34\linewidth}
\centering
\includegraphics[width=1.5in,trim=100 250 50 275,clip]{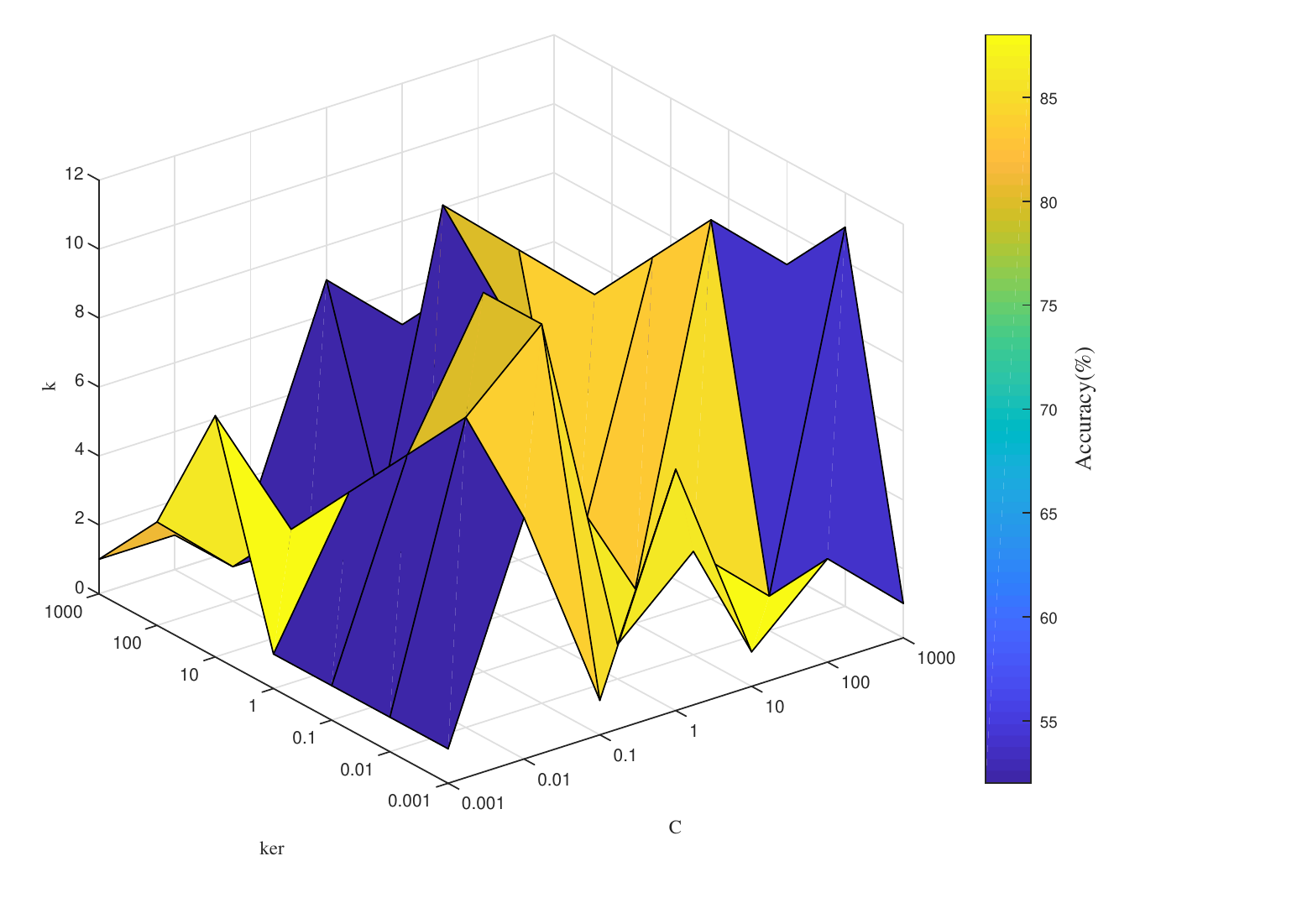}
\end{minipage}%
}%
\subfigure[dal. vs sc.]{
\begin{minipage}[t]{0.34\linewidth}
\centering
\includegraphics[width=1.5in,trim=100 250 50 275,clip]{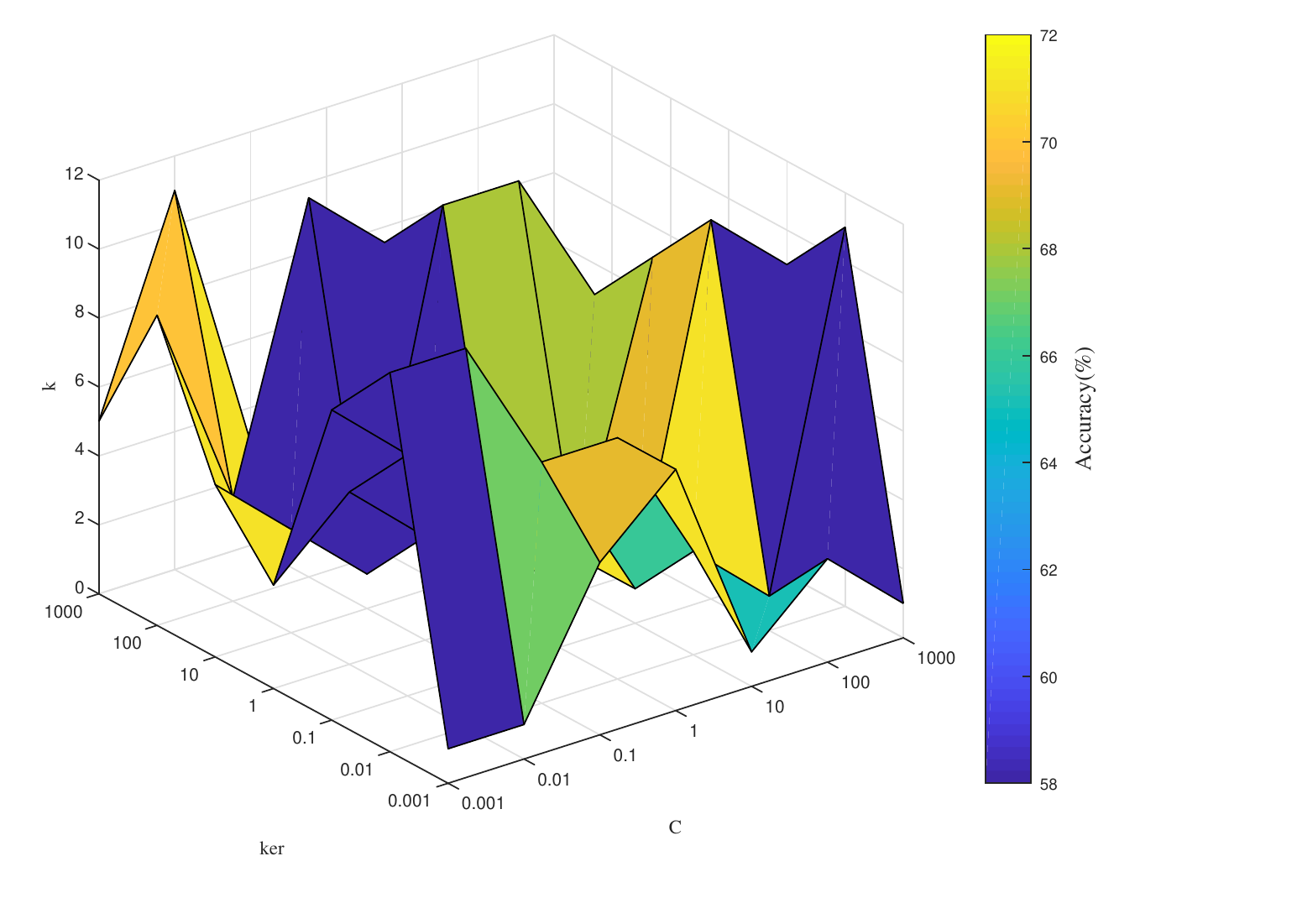}
\end{minipage}
}

\centering
\caption{The figure indicates the highest accuracy k value corresponding to each combination of C and ker on the datasets from AwA. The colors indicate the degree of accuracy.}
\label{canshu}
\end{figure}

The consequences are shown in Fig.\ref{canshu}. It depicts the values of the main parameter k for each combination of parameters C and the kernel parameter corresponding to the highest accuracy rate, in which we control the remaining parameters to be consistent with C to simplify the calculation and graphing. These graphs show that the accuracy of MPWTSVM has something to do with the parameters k, C, $\sigma(ker)$ on all datasets and is sensitive to them all. Therefore, these parameters should be carefully adjusted.

\subsection{Experimental results}
Below, we make the capability of MPWTSVM and all benchmark algorithms under the case of nonlinearity in comparison. Table \ref{result} gives a summary of the detailed information of the overall experimental results on 45 multi-view datasets, including the accuracy, time, and average rank.
`Accuracy' represents the average of five test results, plus or minus the standard deviation.
`Time' means the average training time of five experiments. The bold values in Table \ref{result} demonstrate the best accuracy.

{\setlength{\tabcolsep}{1.0pt} 
{\renewcommand\arraystretch{1.5} 
{
 \scriptsize
\scriptsize
\begin{flushleft}
\begin{longtable}[c]{l|l|l|l|l|l|l|l|l}
\caption{Performance on \emph{AwA} dataset (average accuracy $\pm$ standard deviation of accuracy(\%)).}
\label{result}\\
    \toprule[1.0pt]
	\endfirsthead
	\toprule[1.0pt]
    \multicolumn{9}{l}{continued table \ref{result}}\\
	\midrule
	\endhead
	\bottomrule[1.0pt]
	\endfoot
\textbf{} &
   &
  \textbf{SVM+A} &
  \textbf{SVM+B} &
  \textbf{SVM-2K} &
  \textbf{MVTSVM} &
  \textbf{MCPK} &
  \textbf{PSVM-2V} &
  \textbf{MPWTSVM} \\ \hline
1  & ante. vs. sc.        & 58.50±3.791  & 84.50±3.710          & 68.00±11.374 & 63.50±4.183  & 82.50±6.374          & 84.50±4.108          & \textbf{86.00±6.021} \\
2  & ante. vs. gb.        & 67.00±3.260  & 82.50±7.071          & 75.00±3.062  & 72.50±6.847  & 82.00±4.472          & 85.00±2.500          & \textbf{85.50±4.108} \\
3  & ante. vs. kw.        & 79.00±8.023  & 88.50±5.477          & 83.50±5.184  & 77.00±12.550 & 89.00±3.791          & 88.00±1.118          & \textbf{91.50±3.791} \\
4  & ante. vs. bea.       & 88.50±5.477  & 98.00±2.092          & 92.00±5.420  & 91.00±2.850  & \textbf{98.50±2.236} & 96.50±4.183          & 98.00±1.118          \\
5  & ante. vs. dal.       & 73.50±6.982  & 80.50±6.708          & 73.00±4.809  & 72.50±8.839  & 84.00±7.416          & \textbf{84.00±3.354} & 81.00±7.624          \\
6  & ante.vs. pt.        & 72.50±4.677  & 88.50±8.023          & 75.50±5.701  & 76.00±4.873  & 88.00±3.708          & 87.00±2.092          & \textbf{90.00±5.590} \\
7  & ante. vs. ho.        & 68.00±8.551  & \textbf{84.00±7.416} & 76.00±8.023  & 69.50±7.786  & 83.50±4.183          & 82.00±5.123          & 83.50±3.791          \\
8  & ante. vs. Gs.        & 73.00±9.747  & \textbf{92.50±3.953} & 78.50±17.375 & 74.50±2.739  & 89.00±1.369          & 91.50±4.183          & 92.00±2.739          \\
9  & ante. vs. bw.        & 75.50±4.809  & 84.50±9.585          & 75.50±5.969  & 73.00±5.420  & 83.50±5.184          & 82.00±8.178          & \textbf{84.50±6.471} \\
10 & gb. vs. sc.         & 71.00±10.548 & 87.00±6.225          & 79.50±7.159  & 75.00±6.124  & 86.50±2.850          & 84.50±4.472          & \textbf{88.50±2.850} \\
11 & gb. vs. kw.         & 74.00±9.117  & 82.00±9.906          & 77.00±4.809  & 80.50±7.583  & 84.50±4.809          & 83.00±5.420          & \textbf{85.50±2.092} \\
12 & gb. vs. bea.        & 86.50±4.183  & 91.50±5.755          & 84.00±5.755  & 87.50±6.374  & \textbf{92.50±2.500} & 91.50±4.541          & 92.50±3.536          \\
13 & gb. vs. dal.        & 68.00±8.551  & 76.00±6.755          & 73.00±5.420  & 74.00±7.202  & 76.00±5.184          & 77.00±6.471          & \textbf{79.50±7.159} \\
14 & gb. vs. pt.         & 73.50±8.944  & 85.00±7.071          & 81.00±8.944  & 83.50±4.873  & 86.50±8.768          & 87.50±4.677          & \textbf{90.50±5.969} \\
15 & gb. vs. ho.         & 59.00±8.944  & 70.00±8.292          & 62.50±15.910 & 62.50±7.500  & 65.00±10.155         & 73.50±6.519          & \textbf{76.50±6.755} \\
16 & gb. vs. Gs.         & 76.00±2.236  & 83.00±4.108          & 67.50±8.478  & 76.00±5.755  & 83.50±6.021          & 84.50±6.937          & \textbf{87.50±5.000} \\
17 & gb. vs. bw.         & 68.00±7.374  & 81.00±3.791          & 77.50±3.062  & 77.00±4.809  & 81.50±6.021          & 81.50±6.755          & \textbf{88.50±3.354} \\
18 & kw. vs. sc.         & 80.50±4.809  & 87.00±2.739          & 83.00±6.471  & 81.00±7.202  & 86.00±1.369          & 89.00±3.354          & \textbf{89.50±4.108} \\
19 & kw. vs. bea.        & 81.50±6.519  & 87.00±4.809          & 86.50±6.519  & 85.00±5.590  & 89.00±4.183          & 90.50±2.092          & \textbf{92.00±2.092} \\
20 & kw. vs. dal.        & 68.00±6.708  & 70.50±6.708          & 70.50±4.108  & 72.00±2.092  & 72.00±13.393         & 74.00±3.354          & \textbf{75.50±2.739} \\
21 & kw. vs. pt.         & 74.00±6.275  & 83.50±2.850          & 77.00±4.472  & 71.50±8.944  & 86.00±2.850          & \textbf{86.50±4.183} & 86.00±3.354          \\
22 & kw. vs. ho.         & 70.50±7.159  & 83.50±3.791          & 82.50±8.101  & 76.50±6.021  & 86.00±6.275          & 84.50±7.159          & \textbf{88.00±2.739} \\
23 &
  kw. vs. Gs. &
  69.00±5.184 &
  \textbf{87.50±5.863} &
  74.50±5.701 &
  73.00±3.26 &
  84.50±5.420 &
  85.50±3.260 &
  87.00±5.701 \\
24 & kw. vs. bw.         & 75.50±3.708  & 84.00±6.755          & 78.00±6.225  & 77.00±5.701  & 83.00±8.178          & 82.00±4.809          & \textbf{86.50±2.236} \\
25 & bea. vs. sc.        & 89.50±4.108  & 95.00±4.330          & 88.50±2.236  & 93.00±2.092  & 96.50±1.369          & 95.00±3.536          & \textbf{97.00±1.118} \\
26 & bea. vs. dal.       & 91.50±5.184  & 96.00±2.850          & 95.50±3.708  & 93.00±3.260  & 96.50±2.236          & 95.50±2.092          & \textbf{97.50±3.062} \\
27 &
  bea. vs. pt. &
  88.50±3.354 &
  93.50±2.850 &
  92.50±3.953 &
  90.00±3.062 &
  \textbf{95.50±3.260} &
  93.50±3.791 &
  \textbf{95.50±3.260} \\
28 & bea. vs. ho.        & 88.50±5.184  & 98.00±2.092          & 91.00±5.755  & 91.00±2.236  & 98.00±2.092          & 97.50±2.500          & \textbf{98.50±1.369} \\
29 & bea. vs. Gs.        & 92.00±5.701  & 94.00±6.755          & 94.00±3.354  & 93.00±3.708  & 95.50±3.260          & 96.50±1.369          & \textbf{97.00±2.092} \\
30 & bea. vs. bw.        & 90.50±2.092  & 95.00±4.677          & 89.00±2.236  & 92.00±2.739  & 97.00±1.118          & 96.00±2.85           & \textbf{98.50±2.236} \\
31 & dal. vs. sc.        & 76.00±6.275  & 85.50±7.583          & 74.50±5.969  & 73.50±5.477  & 87.50±6.374          & 85.50±4.809          & \textbf{89.00±2.236} \\
32 & dal. vs. pt.        & 74.00±4.183  & 86.50±3.354          & 79.50±9.906  & 77.00±4.108  & 88.00±2.092          & 89.00±6.275          & \textbf{89.50±4.809} \\
33 & dal. vs. ho.        & 67.00±11.096 & 76.00±5.184          & 76.00±6.755  & 68.00±7.984  & \textbf{82.00±4.472} & 78.50±6.519          & 78.50±6.519          \\
34 & dal. vs. Gs.        & 66.50±5.477  & 84.00±5.755          & 69.50±8.178  & 71.00±6.982  & 81.50±3.791          & 81.50±5.755          & \textbf{85.00±3.953} \\
35 & dal. vs. bw.        & 65.00±5.303  & 78.50±6.021          & 73.50±16.919 & 69.50±11.646 & 84.50±7.159          & \textbf{84.50±3.26}  & 82.00±3.260          \\
36 & pt. vs. sc.         & 60.00±4.677  & \textbf{78.50±3.354} & 65.00±1.768  & 65.00±5.303  & 77.00±7.583          & 76.50±4.541          & 78.00±5.701          \\
37 & pt. vs. ho.         & 70.50±4.809  & \textbf{92.50±3.062} & 87.50±0.000  & 76.50±5.755  & 89.50±4.472          & 92.00±2.739          & 91.50±5.755          \\
38 & pt. vs. Gs.         & 68.00±2.739  & 85.00±6.374          & 81.00±10.093 & 72.00±7.583  & 84.00±4.183          & 82.50±6.124          & \textbf{87.50±5.303} \\
39 & pt. vs. bw.         & 71.50±3.791  & 84.00±7.416          & 75.50±3.708  & 72.00±7.374  & 88.00±3.708          & 87.00±8.551          & \textbf{89.00±4.183} \\
40 & ho. vs. sc.         & 67.00±3.260  & 83.50±4.541          & 76.00±7.202  & 67.50±5.590  & 83.00±3.260          & 80.50±4.472          & \textbf{86.00±4.873} \\
41 &
  ho. vs. Gs. &
  64.00±6.275 &
  \textbf{85.00±3.953} &
  70.50±8.551 &
  71.00±6.755 &
  84.50±9.906 &
  \textbf{85.00±3.953} &
  84.50±5.969 \\
42 & ho. vs. bw.         & 66.50±3.791  & 73.50±7.202          & 68.00±4.108  & 69.00±8.944  & 75.50±8.178          & 75.00±5.000          & \textbf{76.00±4.183} \\
43 & Gs. vs. sc.         & 75.00±8.292  & 90.50±3.708          & 71.00±5.477  & 74.00±6.755  & 90.50±3.260          & 90.50±3.260          & \textbf{93.00±2.092} \\
44 & Gs. vs. bw.         & 65.00±5.303  & 85.50±9.253          & 68.00±7.786  & 70.50±7.159  & 83.50±4.873          & 83.00±4.472          & \textbf{87.50±4.677} \\
45 & bw. vs. sc.         & 64.50±8.178  & 73.00±6.225          & 70.50±7.374  & 72.50±6.124  & \textbf{77.00±9.083} & 73.50±5.477          & 73.50±4.873          \\ \hline
   & \textbf{Avg.Acc.}  & 73.59        & 85.22                & 77.94        & 76.50        & 85.72                & 85.64                & \textbf{87.57}                \\
   & \textbf{Avg.Time.} & 0.111        & 0.138                & 0.347        & 0.067        & 0.088                & 0.858                & \textbf{0.042 }
   \\
   & \textbf{Avg.Rank.} & 6.678        & 3.011                & 5.500        & 5.756    & 2.678               & 2.911               & \textbf{1.467}
    \\
   & *\textbf{W/D/L} & 45/0/0       & 37/2/6               & 45/0/0       & 45/0/0   & 35/5/5               & 38/2/5              & 0/45/0
    \\ \hline
\end{longtable}
*W/D/L is short for Win/Draw/Loss.
\end{flushleft}
\normalsize
 }
 }
 }

The average accuracy of 45 datasets of MPWTSVM is 87.57\%, which is the highest among the seven algorithms followed by MCPK (85.72\%), PSVM-2V (85.64\%), SVM+A (85.22\%), SVM-2K (77.94\%), MVTSVM (76.50\%) and SVM+B (73.59\%). MPWTSVM,  MCPK and PSVM-2V  can follow both the consensus principle and the complementarity principle. Therefore, the three algorithms perform better than the other four. It is worth noting that our model has better generalization performance compared to MCPK and PSVM-2V. The reason is that it utilizes the weighting idea to better exploit the intrinsic connections and differences within and between different perspectives.
SVM-2K and MVTSVM only satisfy the principle of consensus, so the generalization ability is slightly insufficient compared with MPWTSVM, PSVM-2V and MCPK.
Among the several algorithms, the one with the lowest accuracy is SVM+, since the model considers only one perspective and uses the other perspective as its privileged information. 

The average time of MPWTSVM is 0.042s which is the shortest one compared to 
MVTSVM (0.067s), MCPK (0.088s), SVM+A (0.111s), SVM+B (0.138s), SVM-2K (0.347s),  and PSVM-2V (0.858s).
The reason is that the proposed MPWTSVM not only solves two small-scale QPPs, but also uses inter-class weights to remove redundant samples. Therefore, the time complexity can be greatly reduced.

For the same dataset, the accuracy of seven algorithms is ranked and the optimal one is assigned as 1, the sub optimal algorithm is designated as 2, and so on. From this, the average rank of each algorithm in 45 datasets is obtained. Our model has an lowest average rank of 1.467 followed by MCPK(2.678) and PSVM-2V(2.911) among the seven algorithms, which shows its good classification performance.

In addition, Table \ref{result} shows the wins and losses of our algorithm compared to other algorithms. If the accuracy of our MPWTSVM is higher than the other algorithm, it is marked as `Win'; if it is equal to the other one, it is marked as `Draw'; and if it is lower than the other one, it is marked as `Loss'. After comparing our algorithm with other algorithms in 45 data sets, we can count the win-loss situation of each algorithm. It can be seen that compared with SVM+A, SVM-2K, and MVTSVM, our algorithm wins 45 times in 45 data sets; compared with SVM+B, MPWTSVM wins 37 times, loses six times, and draws twice; compared with MCPK, MPWTSVM wins 35 times, loses five times and draws five times; compared with PSVM-2V, MPWTSVM wins 38 times, loses five times and draws twice. It denotes that our algorithm has more prominent advantages compared with other parallel algorithms.

\begin{figure}[ht!]
    \centering          
    \includegraphics[width=11cm]{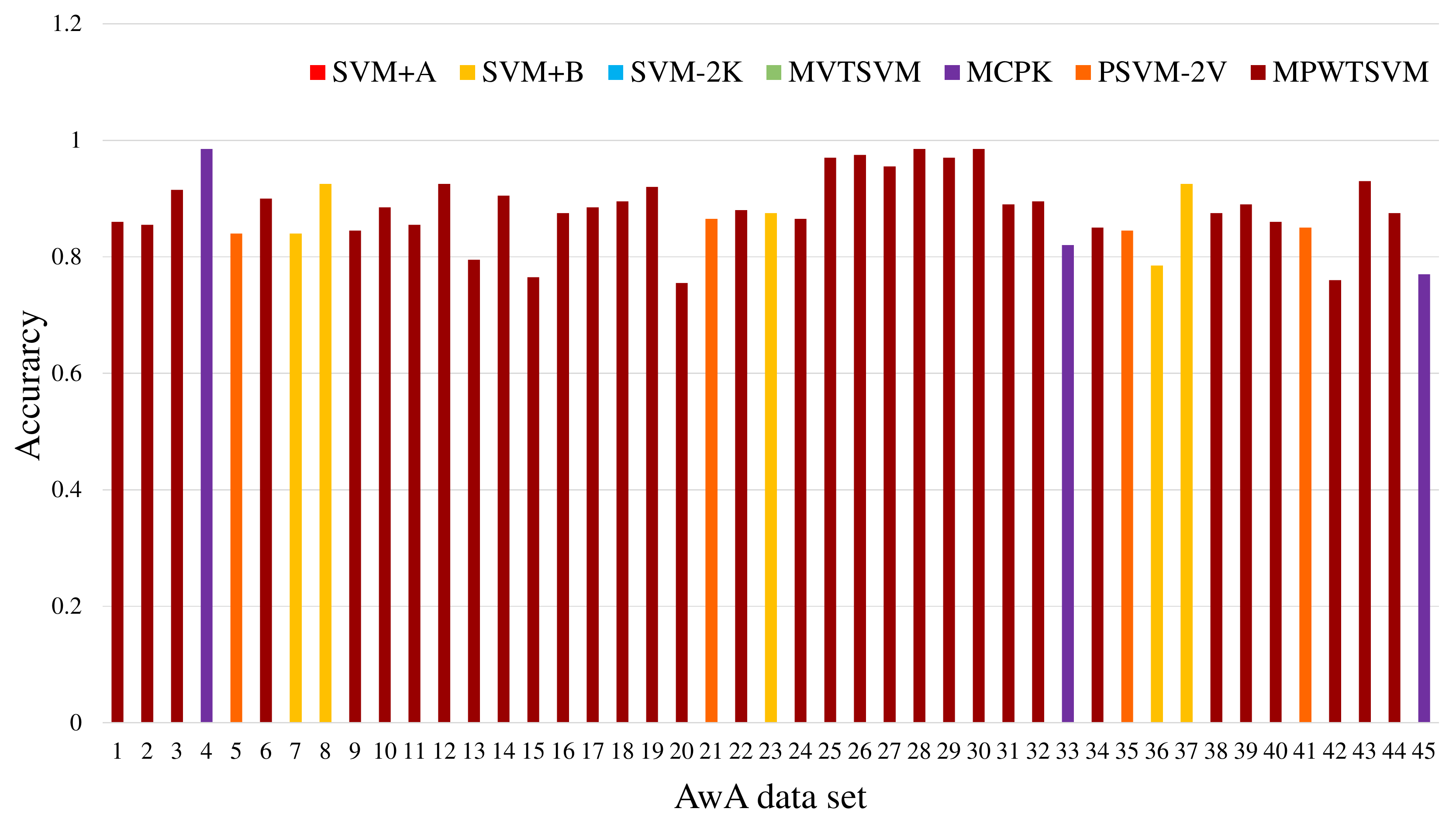}
    \caption{The plot denotes the differences separating the winning method MPWTSVM's (dark red) classification accuracy and the remaining algorithms' average accuracy. }
    \label{accpic}
    \end{figure}

Fig.\ref{accpic} describes the differences separating the winning algorithm's accuracy and the remaining algorithms' average accuracy. The length of each bar reflects the generalization performance of the algorithm, and different colors correspond to different algorithms. We can see that MPWTSVM has a better accuracy among these models. 

\begin{figure}[ht!]
    \centering         
    \includegraphics[width=11.00cm]{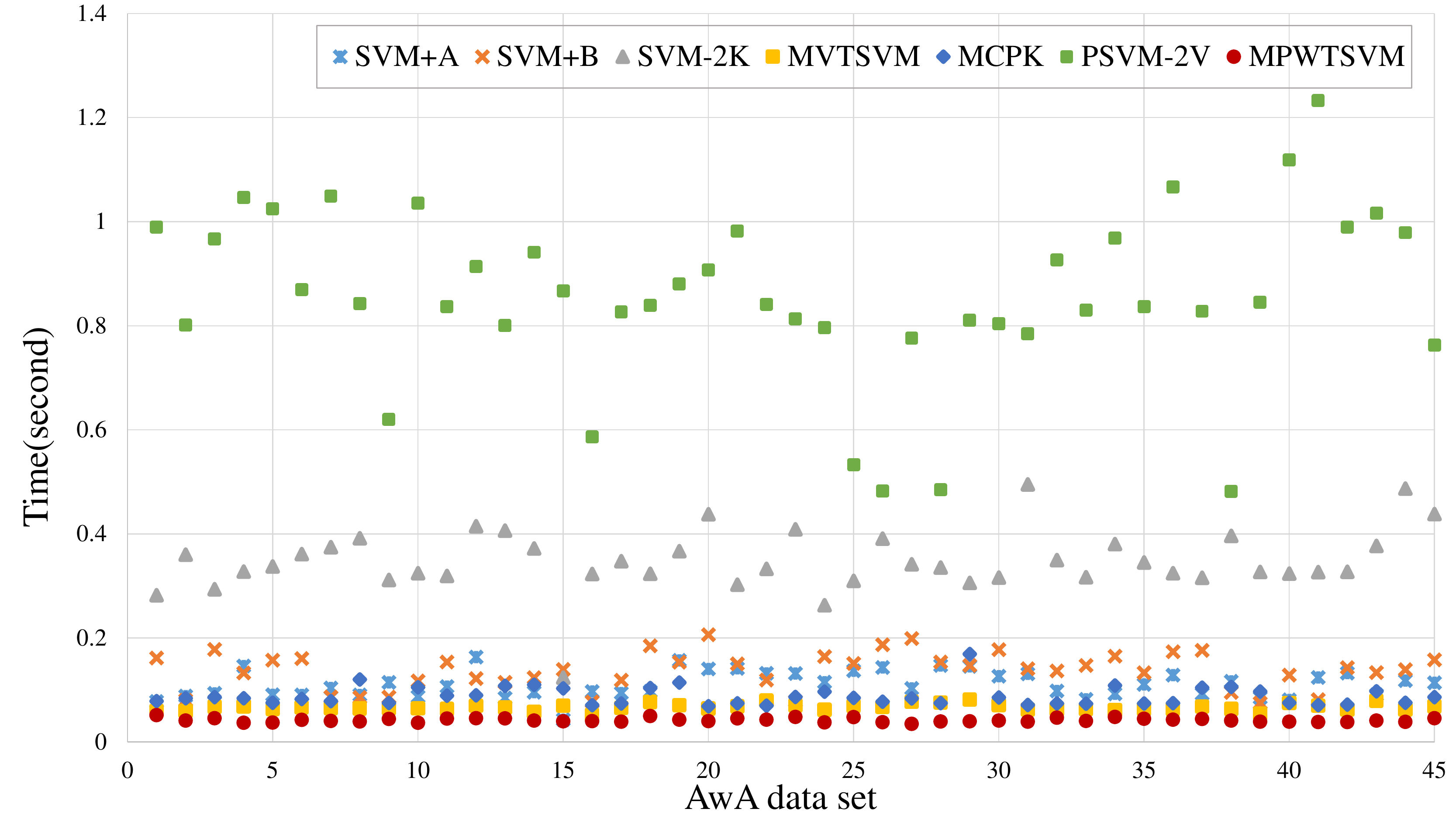}
    \caption{The plot denotes the training time of the following algorithms: SVM+A, SVM+B, SVM-2K, MVTSVM, MCPK, PSVM-2V, and MPWTSVM. }
    \label{timepic}
    \end{figure}

Fig.\ref{timepic} depicts the comparison between the training time of the winning algorithm and the remaining algorithms. The value of the ordinate represents each method's training time. The numbers on the abscissa indicate the number of data sets. It is due to the deletion of redundant samples with the help of the KNN idea, which reduces the running time and improves the computational efficiency. It shows the superior efficiency of MPWTSVM.
These results strengthen the fact that MPWTSVM itself can make full use of intra-class and inter-class information to obtain better classification performance, which means that MPWTSVM has good generalization ability.

\subsection{Friedman test}
We use Friedman test \cite{Friedmantest} for further analysis since the averaged accuracy of our MPWTSVM in Table \ref{result} is not always optimal. For each dataset, the highest accuracy ranking is 1, followed by 2, and so on.
Table \ref{result} also shows the average ranking of the six comparison algorithms. As can be seen from the Avg.Rank line in Table \ref{result}, the average ranking of MPWTSVM is 1.467, which ranks the bottom of the six methods. The outcome shows that MPWTSVM proposed by us has the best performance among the six comparison methods.

The following null hypothesis is made: the 7 methods are identical.
Friedman statistics can be computed using the following formula:
\begin{equation} \label{Friedman}
\chi^2_F=\frac{12N}{k(k+1)} \left[\sum\limits_j R^2_j-\frac{k(k+1)^2}{4}\right]
\end{equation}
where $R_j=\frac{1}{N} \sum\limits_i r_i^j$ and $r_i^j$ is the $j$th of $k$ algorithms on the $i$th of N datasets.
From (\ref{Friedman}) and with the help of R language, we can get $\chi^2_F$ is 222.11 and the P-value of the hypothesis test is $2.2\times10^{-16}$.
If $\alpha = 0.05$, P-value is much less than $\alpha$. This means that it denies the null hypothesis above. In other words, the differences between these seven methods are obvious.

\begin{figure}[ht!]
    \centering
    \includegraphics[width=11cm]{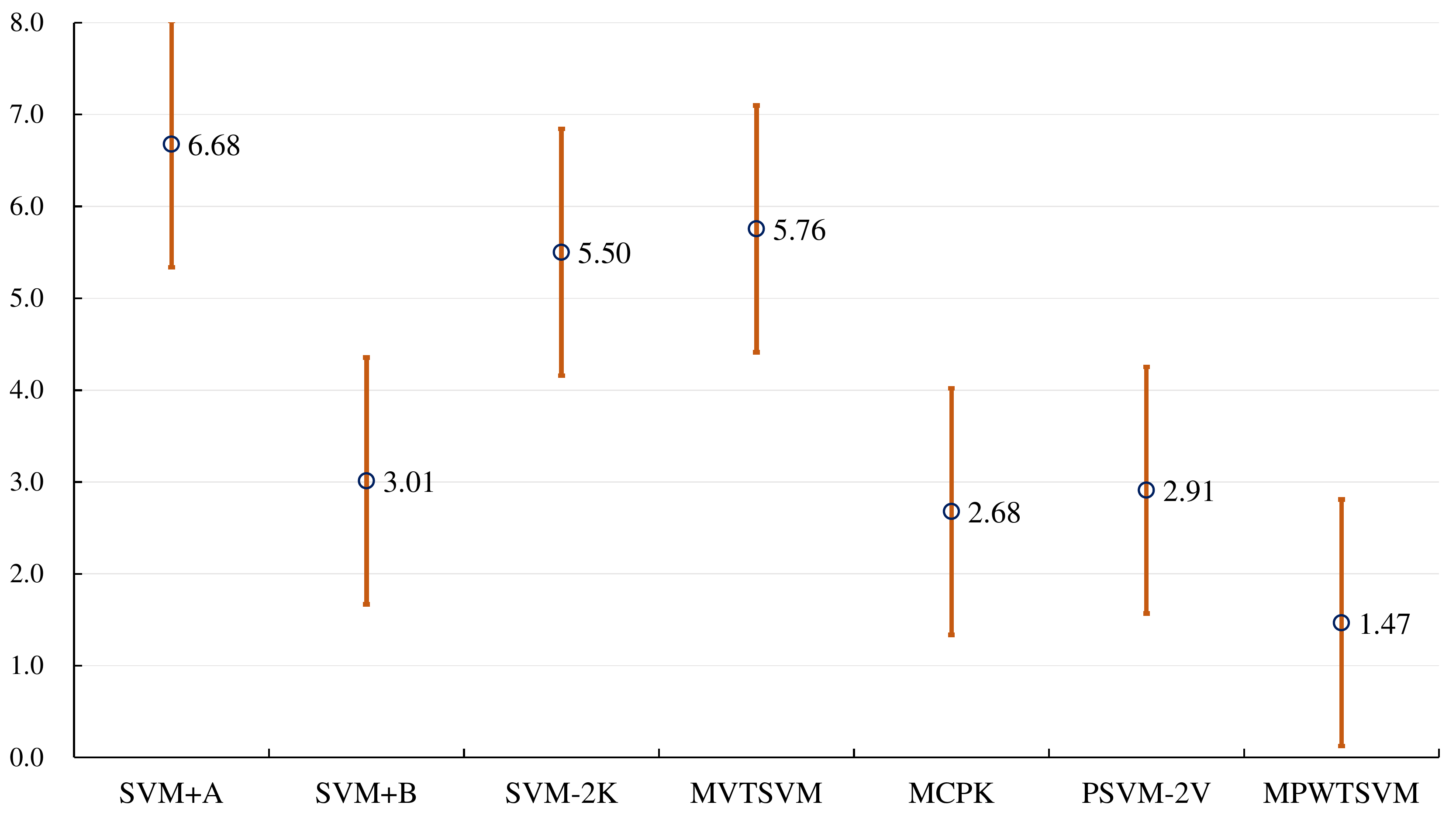}
    \caption{Friedman test.}
    \label{Friedpic}
    \end{figure}

Since the null hypothesis is rejected, Nemenyi test \cite{nemenyi1963} can be further conducted.
The hollow circle represents the average ranking of the seven algorithms, and the critical difference CD is shown in the straight lines centered on "$\circ$"
in Fig.\ref{Friedpic}, where 
\begin{equation} \label{Nemenyi}
    CD=q_\alpha \sqrt{\frac{k(k+1)}{6N}}.
\end{equation}
We can get
$q_\alpha = 2.949$ and $CD = 1.343 $ under the significance level $\alpha = 0.05$ by calculation.
The consequence in Fig.\ref{Friedpic}  shows that the proposed MPWTSVM has remarkably better performance than the other models at the confidence level of $95\%$ .

\section{Conclusions}
In this paper, we propose a weighted MVL kernel method based on privileged information termed MPWTSVM, which satisfies the principle of consensus and complementarity at the same time. The consensus principle is ensured by minimizing the coupling items of the two views in the original goal. We implement the principle of complementarity by establishing a privileged information paradigm and by drawing on multi-view learning. In the two-classification process, we weigh the samples from two perspectives to obtain the weights of different types and the same type in each view. The internal similarity information of samples of the same category is captured. During the learning process, our model entirely integrates the information of diverse views and maintains the characteristics of diverse views to a certain extent. Therefore, the proposed MPWTSVM has better classification accuracy and faster solving speed. We use the standard QP solver to solve MPWTSVM and compare it with the other five algorithms. Through numerical experiments on 45 sets of multi-view binary datasets demonstrate the validity and efficiency of our MPWTSVM. In the future, we will extend MPWTSVM to the case of multiple (more than two) views under the guideline of two principles. Other effective optimization algorithms like the alternating direction methods of multipliers(ADMM) and other solution ways can also be put into consideration before long.

\section*{Acknowledgments}
This work was supported in part by the Fundamental Research Funds for the Central Universities (No. BLX201928).

\bibliography{reference.bib}   
\end{document}